\documentclass[preprint,3p,12pt]{elsarticle}

\usepackage{amssymb}

\usepackage{booktabs}
\usepackage{rotating}
\usepackage{placeins}
\usepackage{xspace}
\usepackage{multicol}
\usepackage{subcaption}
\usepackage{color}
\usepackage[table]{xcolor}
\usepackage{adjustbox}
\usepackage{xcolor}
\usepackage{soul}
\usepackage{makecell}
\sethlcolor{yellow} 
\def\XS{\xspace}
\def\figureabvr{Figure\XS} 
\def\figuresabvr{Figures\XS} 
\usepackage{url}
\usepackage{hyperref}

\usepackage{amssymb}
\usepackage{pifont}
\newcommand{\xmark}{\ding{55}}%

\usepackage{longtable}
\usepackage{pdflscape}
\usepackage{multirow}

\def\tableabvr{Table\XS}
\def\eg{\textit{e.g.,}\XS}
\def\etal{\textit{et al.}\XS}

\def\ie{\textit{i.e.,}\xspace}
\definecolor{lightpink}{HTML}{fde9a9}
\definecolor{lightblue}{HTML}{80B3B3}
\definecolor{olive}{HTML}{B3B34D}
\definecolor{pink}{HTML}{F5B3D1}
\definecolor{lightgreen}{HTML}{D9F7B3}
\definecolor{rose}{HTML}{ffd699}

\newcommand\tabitem{\makebox[1em][r]{\textbullet~}}

\newif\ifrevisions
\revisionsfalse 

\newcommand{\rev}[1]{%
  \ifrevisions
    \textcolor{red}{#1}%
  \else
    #1%
  \fi
}
\newcommand{\revbox}[1]{%
  \ifrevisions
    \fcolorbox{red}{white}{#1}%
  \else
    #1%
  \fi
}

\newcommand{\revtable}[1]{%
  \ifrevisions
    {\color{red}#1}
  \else
    #1%
  \fi
}

\journal{Medical Image Analysis}

\begin{document}

\begin{frontmatter}

\title{Detecting Dental Landmarks from Intraoral 3D Scans: the 3DTeethLand challenge}

\author[crns]{Achraf Ben-Hamadou\corref{correspondingauthor}}
\ead{achraf.benhamadou@crns.rnrt.tn}
\author[crns]{Nour Neifar}
\author[crns]{Ahmed Rekik}
\author[udini]{Oussama Smaoui}

\author[udini]{Firas Bouzguenda}

\author[inria]{Sergi Pujades}
\address[crns]{Centre de Recherche en Num\'{e}rique de Sfax,  Laboratory of Signals, Systems, Artificial Intelligence and Networks, Technop\^{o}le de Sfax, 3021 Sfax, Tunisia}
\address[udini]{Udini, 37 BD Aristide Briand, 13100 Aix-En-Provence, France}
\address[inria]{Inria, Univ. Grenoble Alpes, CNRS, Grenoble INP, LJK, France}

\author[inst1]{Niels {van Nistelrooij}}
\author[inst1]{Shankeeth Vinayahalingam}

\affiliation[inst1]{Department of Oral and Maxillofacial Surgery, Radboud University Medical Center, Geert Grooteplein Zuid 10, 6525 GA Nijmegen, Netherlands}

\author[inst5]{Kaibo Shi}
\author[inst5]{Hairong Jin}
\author[inst5]{Youyi Zheng}

\affiliation[inst5]{organization={State Key Lab of CAD\&CG, Zhejiang University, Hangzhou, 310058}, country={China}}

\author[inst3,inst4]{Tibor Kubík}
\author[inst3]{Oldřich Kodym}
\author[inst3,inst4]{Petr Šilling}
\author[inst3]{Kateřina Trávníčková}
\author[inst3]{Tomáš Mojžiš}
\author[inst3]{Jan Matula}

\affiliation[inst3]{organization={TESCAN 3DIM, s.r.o.}, 
            city={Brno}, 
            country={Czech Republic}}

\affiliation[inst4]{organization={Department of Computer Graphics and Multimedia, Brno University of Technology}, 
            city={Brno}, 
            country={Czech Republic}
            }

\author[inst6]{Jeffry Hartanto}

 \author[inst7]{Xiaoying Zhu}

\author[inst8]{Kim-Ngan Nguyen}
\author[inst9]{Tudor Dascalu}

\affiliation[inst6]{text={National Dental Centre Singapore} } 

\affiliation[inst7]{organization={Guangxi Colleges and Universities Key Laboratory of Intelligent Software},country={ Wuzhou University}
}

\affiliation[inst8]{text={National University of Singapore}}
\affiliation[inst9]{organization={Department of Computer Science} , country = {University of Copenhagen}}

\author[inst10]{Huikai Wu}
\affiliation[inst10]{text={Hangzhou ChohoTech Inc}}

\author[inst11]{Weijie Liu}
\author[inst11]{Shaojie Zhuang}
\author[inst11]{Guangshun Wei}
\author[inst11]{Yuanfeng Zhou}

\affiliation[inst11]{organization={School of Software, Shandong University}, 
 country={China}
}

\begin{abstract}
Teeth landmark detection is a \rev{key} task in modern orthodontics\rev{, supporting advanced diagnosis, personalized treatment planning, and effective monitoring of treatment progress.} However, several significant challenges may arise due to the intricate geometry of individual teeth and the substantial variations observed across different individuals. To address these complexities, the development of advanced techniques, especially through the application of deep learning, is essential for the precise and reliable detection of 3D tooth landmarks. In this context, the 3DTeethLand challenge was held in \rev{conjunction} with the International Conference on Medical Image Computing and Computer-Assisted Intervention (MICCAI) in 2024, calling for algorithms focused on teeth landmark detection from intraoral 3D scans.
\rev{This challenge introduced a publicly available dataset for 3D dental landmark detection from 340 intraoral scans, providing a standardized benchmark to evaluate state-of-the-art approaches and encouraging methodological advances toward addressing this clinically problem.} \rev{A total of 49 teams participated, and 6 teams reached the final phase. The winning team achieved a rank score of 0.91, with a mean Average Precision of 0.78 and a mean Average Recall of 0.65, demonstrating a balance between precision and recall.} \rev{Top teams achieved high precision with different strategies: the first-ranked team used a two-stage Stratified Transformer with segmentation and weighted DBSCAN, while the second-ranked team adopted a single-stage DGCNN with offset regression and class-specific non-maximum suppression.}
\end{abstract}


\end{frontmatter}

\section{Introduction}
\label{sec:introduction}

Digital tools like computer-aided design (CAD) have become essential in contemporary dentistry, enabling highly precise treatment planning. Advanced intraoral scanners (IOSs) are particularly important in orthodontic CAD systems, as they produce accurate 3D digital models of the teeth \cite{angelone2023diagnostic}. These digital models serve as the foundation for various orthodontic applications, such as teeth alignment and treatment simulation \cite{logozzo2014recent,eggmann2024recent}. One of the most essential tools for such applications is the detection of key dental landmarks. These landmarks, including features such as cusps and mesial-distal locations, serve as fundamental reference points for diagnosing dental conditions, designing personalized treatment plans, and evaluating treatment progress \cite{woodsend2021automatic}. Accurate identification of these landmarks plays an essential role in advancing orthodontic procedures, enabling precise alignments, improving occlusal relationships, and enhancing the overall efficiency and outcomes of dental care. However, several significant challenges may arise due to the intricate geometry of individual teeth, including variations in shape, size, orientation, as well as the presence of pathological teeth \cite{palone2020analysis}. Considerable inter-individual differences, such as variations in dental arch forms, tooth alignment, and occlusal relationships, further complicate these challenges. 

To address this, developing advanced techniques has become a central focus of research, particularly by leveraging the potential of deep learning for the accurate detection of 3D tooth landmarks. However, the limited availability of publicly accessible data presents a significant challenge, as the success of such approaches, as well as their fair comparison, depends heavily on the availability of well-annotated datasets.  

In this context, the 3DTeethLand challenge 
\footnote{\href{https://crns-smartvision.github.io/teeth3ds/\#3DTeethLandSec}{https://crns-smartvision.github.io/teeth3ds/\#3DTeethLandSec}} was organized in collaboration with the International Conference on Medical Image Computing and Computer-Assisted Intervention (MICCAI) in 2024, alongside two other challenges: ToothFairy \cite{bolelli2024segmenting} and STS \cite{wang2025miccai}.

The 3DTeethLand challenge specifically called for algorithms aimed at detecting teeth landmarks from intraoral 3D scans. To support this effort, we also present the first publicly available database on intraoral scans prepared for the challenge, to advance research in 3D tooth landmark detection.  \rev{The novel dataset enriches the 3D intraoral scans from our previous 3DTeethSeg'22 challenge with newly annotated dental landmarks and an extended set of tasks: namely, tooth detection, segmentation, labeling, and landmark localization.} The result announcement was held during the satellite event dedicated to the challenge at MICCAI 2024.

\rev{As a brief preview of the main challenge outcomes, the winning team achieved a rank score of 0.91, with a mean Average Precision of 0.78 and a mean Average Recall of 0.65. Their main contribution was a two-stage Stratified Transformer pipeline that combined full-arch segmentation, high-resolution per-tooth landmark detection, and weighted DBSCAN clustering. The second-ranked team achieved a rank score of 0.83, mAP of 0.77, and mAR of 0.63, using a single-stage DGCNN-based offset regression framework with point-wise refinement and class-specific non-maximum suppression. Beyond these quantitative results, the challenge provided several important insights: notably, while individual tooth segmentation is a common initial strategy, it introduces a significant computational burden and is not strictly necessary to achieve high accuracy. Furthermore, participants demonstrated that dense local 3D sampling is necessary, though not sufficient, for precise landmark localization. Finally, targeted post-processing strategies, such as class-specific non-maximum suppression, were shown to consistently improve performance.}

\section{\rev{Related work}}
\label{sec:relatedwork}
\rev{Unlike the segmentation task, which is widely studied \cite{ben20233dteethseg,rekik2025tseglab}, the number of studies on tooth landmark detection (\ie localization) is still limited. Existing methods can be broadly categorized into three categories: traditional geometry-based methods, deep learning-based direct prediction methods, and hybrid coarse-to-fine frameworks. Despite recent advances, most approaches remain limited by dataset size, annotation inconsistency, and lack of robustness to clinical variability, motivating the need for standardized benchmarks such as the present challenge.}

\rev{\textbf{Geometry-based Landmark Detection Methods:} Early approaches relied on handcrafted geometric features derived from dental surface morphology. These methods typically exploit curvature peaks, surface orientation, and rule-based anatomical heuristics. For example, Woodsend \etal \cite{woodsend2021automatic} proposed a multi-step pipeline that estimates dental surface orientation, detects local geometric maxima for landmark localization, performs tooth surface segmentation, and applies rule-based filtering combined with jaw-line fitting for anatomical labeling. Similarly, \cite{triarjo2023automatic} combined deep learning-based segmentation with weighted mesh point voting strategies for landmark detection, inspired by geometric orientation cues.}

\rev{While these methods provide interpretable geometric priors, they suffer from limited generalization capability. Handcrafted descriptors are highly sensitive to acquisition noise, inter-patient anatomical variability, and pathological cases such as missing teeth or jaw deformities. Consequently, purely geometry-driven pipelines are difficult to deploy in large-scale clinical settings.}

\rev{\textbf{Deep Learning-based Dense Representation Methods:}
To overcome the limitations of handcrafted features, recent works have explored deep neural networks to learn discriminative surface representations. For instance, Wei \etal \cite{wei2022dense} proposed encoding landmarks and tooth axes as dense fields defined over tooth surfaces. Their framework leverages multi-scale feature extraction modules to capture both local geometric details and global structural context. However, this method presents several limitations. It is trained on point clouds of complete, isolated teeth, which reduces its robustness when dealing with partially damaged or incomplete tooth structures. The method is also sensitive to input data quality and requires preprocessed point clouds extracted from CBCT scans, limiting its generalization and clinical applicability.}

\rev{More generally, while dense field-based approaches provide strong representation capacity for modeling complex dental surface geometry, they often lack explicit anatomical or structural constraints. As a result, these methods may produce spatially inconsistent landmark predictions, particularly under challenging acquisition conditions such as noisy scans, partial observations, or anatomical abnormalities. This limitation highlights the trade-off between representation flexibility and anatomical reliability in purely data-driven landmark detection frameworks.}

\rev{\textbf{Hybrid Segmentation-to-Landmark Detection Frameworks:} A major research trend has been the development of hybrid frameworks combining tooth segmentation with subsequent landmark localization. Wu \etal \cite{wu2022two} proposed the TS-MDL framework, which adopts a two-stage strategy. First, an iMeshSegNet \cite{lian2020deep} is used to perform tooth-level segmentation on downsampled meshes. Then, a PointNet-based regressor is applied to tooth-specific regions of interest to predict heatmaps encoding the landmark locations. While TS-MDL offers high computational efficiency by utilizing heatmap and field-based regressions, a critical limitation of this architecture is its reliance on a predefined number of landmarks specified during the clustering stage, which makes the method unsuitable for clinical cases involving anatomical abnormalities, missing cusps, or partially damaged teeth where the number and type of landmarks deviate from the model's fixed training assumptions. Similarly, DLLNet \cite{lang2021dllnet} introduced a coarse-to-fine architecture for localizing 68 anatomical landmarks. The method first performs tooth segmentation using MeshSegNet, followed by patch-based refinement using multi-task feature learning. The model integrates curvature descriptors with spatial coordinates and normal vectors, and uses attention mechanisms to improve feature selection during localization. However, this method remains highly dependent on accurate tooth segmentation and complete anatomical structures. Segmentation errors or anatomical anomalies, such as missing teeth, propagate through the coarse-to-fine pipeline and significantly degrade landmark localization performance, limiting robustness in real clinical scenarios.}

\rev{Although hybrid pipelines generally achieve higher accuracy than geometry-based methods, they remain vulnerable to error propagation from the segmentation stage. In particular, inaccurate tooth partitioning directly degrades landmark prediction quality. Moreover, these approaches typically assume a fixed number of anatomical landmarks, limiting their robustness in cases involving missing teeth or other anatomical variations.}

\rev{\textbf{Direct Landmark Detection and Detection-style Formulations:} Recent works reformulate landmark localization as an object detection or regression problem. For example, DentalPointNet \cite{lang2022dentalpointnet} adopts a two-stage detection pipeline inspired by region proposal networks. The first stage performs coarse landmark localization using curvature-based region proposals, while the second stage refines predictions using local patch-based feature extraction. By treating landmarks as detection targets, this formulation reduces false negative rates in difficult regions. However, similar to other deep learning pipelines, performance remains dependent on training data diversity and annotation quality. Additionally, detection-based formulations still struggle to enforce global anatomical relationships between landmarks.}

\rev{Departing from direct 3D mesh-based learning strategies, Tibor \etal \cite{Tibor2022} reformulated dental landmark detection as a multi-view 2D inference problem. Rather than modeling geometric relationships explicitly in 3D space, the approach renders the surface from multiple viewpoints and performs heatmap regression using an Attention U-Net, followed by 3D reconstruction via RANSAC consensus. To address cases involving missing teeth, the method further incorporates an uncertainty estimation scheme that combines maximum heatmap activation confidence with multi-view consistency measures. This paradigm trades explicit geometric reasoning for multi-view projection consistency, introducing a fundamentally different balance between computational efficiency and geometric fidelity. Nevertheless, the method remains sensitive to mesh alignment and rotational variations, requiring manual correction when alignment errors are large. Furthermore, it exhibits limited robustness to severe anatomical shifts or rare tooth configurations (\eg third molars), and faces a practical trade-off between inference time and marginal accuracy gains as the number of viewpoints increases.}

\rev{\textbf{Summary and Research Gap:} Despite significant progress in dental landmark detection through geometry-based, hybrid, and deep learning-driven frameworks, several fundamental challenges remain unresolved. Current methods often lack robustness to anatomical variability, including missing or partially damaged teeth, severe crowding, and rare structures such as third molars. Many architectures rely on predefined landmark configurations or accurate intermediate segmentation, making them vulnerable to error propagation and limiting adaptability in atypical clinical scenarios.}

\rev{Furthermore, most approaches depend primarily on local geometric cues without incorporating explicit anatomical or topological constraints, which may result in spatial inconsistencies when anatomical configurations deviate from training assumptions. Performance also remains uneven across landmark types, particularly in geometrically smooth regions where supervision signals are inherently ambiguous. Finally, existing methods are typically evaluated on limited and homogeneous datasets, with heterogeneous annotation standards and acquisition protocols, hindering fair comparison and reproducibility.}

\rev{These limitations highlight the need for a standardized and comprehensive benchmark capable of assessing landmark detection performance under realistic clinical variability, diverse acquisition settings, and consistent evaluation criteria.}
\section{Challenge setup}
\label{sec:setup}
This section provides an overview of our database, including details on data annotation and recorded information.

\begin{figure}[h!]
    \centering

    \begin{subfigure}{0.25\textwidth}
        \includegraphics[width=\textwidth]{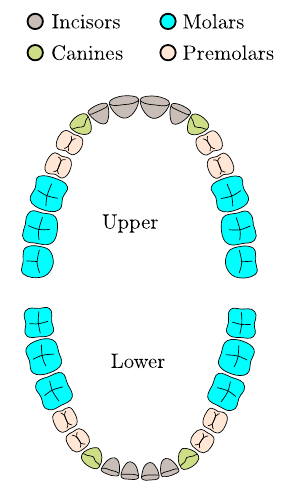}
        \caption{Dental arch structure.}
        \label{fig:teeth_structure}
    \end{subfigure}
    \hspace{15mm}
    \begin{subfigure}{0.5\textwidth}
        \includegraphics[width=\textwidth]{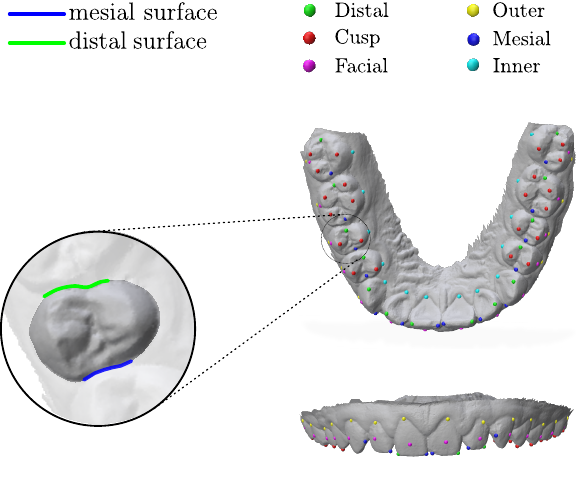}
        \caption{Teeth landmarks.}
        \label{fig:teeth_landmarks}
    \end{subfigure}

    \caption{Overview of tooth structure and anatomical landmarks. 
    (a) Illustration of the human dental arches showing the distribution 
    of incisors, canines, premolars, and molars in the upper and lower jaws. 
    (b) 3D intraoral scans annotated with key dental landmarks, including 
    distal, mesial, facial, cusp, outer, and inner points used for geometric analysis.}
    \label{fig:dental_arch_structure}
\end{figure}
\subsection{Database}

\rev{Dental landmark detection has been explored using multiple imaging modalities, including intraoral scans, cone-beam computed tomography (CBCT), and cephalometric radiographs. However, the majority of these datasets are developed in-house and are not publicly available \cite{lang2022dentalpointnet,Tibor2022,dai2024cone}. Among the few publicly accessible resources, some datasets are based on volumetric CT scans but are primarily designed for 3D craniofacial or skull landmark detection rather than dental analysis. For example, one such CT-based dataset \cite{he2024anchor} includes only a limited subset of dental landmarks within broader craniofacial tasks. Other publicly available datasets rely on 2D cephalometric radiographs, such as the Aariz dataset \cite{khalid2025benchmark}, and the dataset proposed in the Automated Detection and Analysis for Diagnosis in Cephalometric X-ray Image challenge \cite{wang2016benchmark}, which include only a small number of dental landmarks within more general anatomical annotations. These public datasets remain limited in anatomical coverage and are not specifically designed for dental analysis. In contrast, our dataset provides a more comprehensive representation of dental anatomy and associated tasks. To our knowledge, our 3DTeethLand challenge introduces the first publicly available dataset \footnote{\href{https://crns-smartvision.github.io/teeth3ds/}{https://crns-smartvision.github.io/teeth3ds/}} for dental landmark detection from intraoral 3D scans}, which has the potential to accelerate progress in this field. \rev{Beyond their diagnostic and analytical value, intraoral 3D scanning technologies also offer several clinical advantages, including their radiation-free and non-invasive nature, reduced patient discomfort, and simplified acquisition \cite{suese2020progress}.}

The database \cite{ben2022teeth3ds+} was initially created for the 3DTeethSeg'22 MICCAI challenge \cite{ben20233dteethseg}, to advance research in the analysis of intraoral 3D scans on the tasks of teeth identification, segmentation, and labeling. The dataset was collected following the European General Data Protection Regulation (GDPR), in collaboration with orthodontists and dental surgeons with over five years of professional experience from partner dental clinics, primarily based in France and Belgium. All patient information has been anonymized to ensure complete confidentiality. For each patient in the dataset, two 3D scans were obtained, one for the upper jaw and one for the lower jaw. The dataset is released under the CC BY-NC-ND license, allowing sharing with attribution while restricting commercial use and derivative works.

For the 3DTeethLand Challenge, we provided landmark annotations on a dataset comprising 340 intraoral scans (IOS). The dataset is split into two groups:
\begin{itemize}
    \item The first group includes 240 scans from the Teeth3DS dataset, including segmentation and labeling annotations, and serving as the training set for the challenge.
    \item The second group consists of 100 scans \rev{used as the hidden private test set during the challenge.}
\end{itemize}

\rev{During the challenge, the test set annotations were kept hidden to ensure a fair evaluation of the submitted methods. After the completion of the challenge, the test set and its corresponding landmark annotations were made publicly available together with the dataset\footnote{\href{https://osf.io/gvjah/overview}{https://osf.io/gvjah/overview}}, allowing future researchers to benchmark their methods against the reported results. In addition, the evaluation scripts used during the challenge are publicly available in the project repository\footnote{\href{https://github.com/crns-smartvision/3DTeethLand}{https://github.com/crns-smartvision/3DTeethLand}}, together with a Docker-based evaluation pipeline to facilitate reproducible evaluation.}

\subsection{Data acquisition:}

The scans were acquired using state-of-the-art intraoral 3D scanners—specifically the Primescan (Dentsply), TRIOS 3 (3Shape), and iTero Element 2 Plus. These devices are widely adopted in clinical practice and are known to produce high-quality 3D models, with accuracies ranging from 10 to 90 µm and point resolutions between 30 and 80 pts/mm².

No specific acquisition protocol was imposed other than requiring complete coverage of both the upper and lower jaws. Acquisitions were performed by orthodontists or dental surgeons with more than five years of professional experience, while the challenge's clinical evaluators have over ten years of expertise in orthodontics, dental surgery, and endodontics.

The dataset includes patients undergoing either orthodontic treatment (50\%) or prosthetic treatment (50\%). Ethical approval for data collection and use was granted by the Local Ethics Committee of the Digital Research Center, University of Sfax (Tunisia) on 28 October 2020 under approval number PV-CS-28/10/2020. In addition, the study was validated by DELSOL Avocats (Paris, France), a law firm specializing in GDPR compliance.

\subsection{Data annotation}
Teeth have a complex anatomical structure, consisting of various surfaces and features. They are categorized into four main types based on their shape and function, as shown in \figureabvr \ref{fig:teeth_structure}: incisors, canines, premolars, and molars. Landmarks on these teeth serve as key reference points, facilitating the accurate measurement of tooth orientations, spacing, and occlusion. To ensure precise analysis of tooth positioning and alignment, specific dental landmarks are identified on each tooth, providing a standardized approach for consistent and reproducible assessments. 

Annotators were instructed to identify and label every tooth landmark class on each tooth within the intraoral 3D scans.
The annotation process was carried out iteratively by professional annotators using a custom tool developed by our clinical partner, Udini\rev{, which is publicly accessible}\footnote{\rev{\url{https://seglab.udini.ai}}}\rev{, and can be used both for landmark annotation and inference. To avoid introducing bias in the annotation process, all landmarks were annotated manually from scratch, without any AI-based prediction or assistance. 
The tool provides a 3D viewer with standard interaction functionalities allowing annotators to add, adjust, and remove landmarks on the intraoral scans (see Appendix \figureabvr~\ref{fig:main_interface_annotation_tool} for a screenshot of the tool's main interface). On average, annotating a single scan took 7 minutes and 16 seconds. The standard deviation was 1 minute 55 seconds, with a minimum of 4 minutes 6 seconds and a maximum of 13 minutes 12 seconds, reflecting the variability in scan complexity and landmark density.
}  

Following this, a final validation phase was carried out by experienced dentists. Each scan was initially annotated by a single annotator, after which all annotations were systematically reviewed and verified by the three clinical evaluators of the challenge. This multi-step procedure helps ensure the accuracy and consistency of the landmark annotations.

\figureabvr\ref{fig:teeth_landmarks} depicts an overview of the landmarks annotation. This annotation includes: 

\paragraph{Mesial and Distal points}
The mesial surface of a tooth refers to the side of the tooth that faces the center of the mouth. While the distal surface is the opposite of the mesial, meaning it faces away from the midline of the mouth (see \figureabvr\ref{fig:teeth_landmarks}). The mesial and distal points represented in blue and green, respectively, in \figureabvr\ref{fig:teeth_landmarks} are located on the mesial and distal surfaces. These points are crucial for assessing the alignment and positioning of the tooth within the dental arch, helping to evaluate its relationship to adjacent teeth.

\paragraph{Cusp points}
The cusp refers to the pointed or rounded elevation on the chewing surface of a tooth. The cusp point, highlighted in red in \figureabvr\ref{fig:teeth_landmarks}, is situated at the highest point of the cusp, playing a key role in understanding the occlusion relationship between the upper and lower teeth during biting and chewing. The number of cusp points varies depending on the type of tooth. For instance, premolars typically have two cusps, whereas molars typically present three to five cusps (see \figureabvr\ref{fig:teeth_landmarks}).

\paragraph{Inner and outer points}
The inner points, represented in cyan in \figureabvr\ref{fig:teeth_landmarks}, and outer points, depicted in yellow, are defined at the boundary where the tooth meets the gingiva, serving as key anatomical reference points. The inner point is positioned on the lingual or palatal side of the tooth, closer to the tongue or the palate, while the outer point is located on the buccal side, facing the cheeks or lips. These landmarks are essential for analyzing the spatial pose and orientation of the tooth within the dental arch.

\paragraph{Facial axis point} The facial axis point, represented in magenta, is located at the center of the facial surface of each tooth, which is the side visible from the front of the mouth. 
This point is significant for understanding the tooth's angulation and inclination and is crucial for accurate alignment analysis and orthodontic planning.

The distribution of the test landmarks across six categories is as follows:

\begin{multicols}{2}

\begin{itemize}
    \item Cusp: 2343 points
    \item Mesial: 1374 points
    \item Inner: 1351 points

    \item Facial: 1358 points
    \item Distal: 1366 points
    \item Outer: 1350 points

\end{itemize}
\end{multicols}

\rev{\tableabvr \ref{tab:dataset_statistics} provides a statistical summary of the dataset, including the number of patients, scans, and annotated landmarks in the training and testing sets for both upper and lower jaws, as well as detailed statistics on the data sources.}

\begin{table}[h!]
\begin{center}
\caption{\label{tab:dataset_statistics} \rev{Dataset statistics for training and testing splits.}}
\setlength{\tabcolsep}{10pt} 
\renewcommand{\arraystretch}{1}

\rev{
\begin{tabular}{llccc}
\toprule
 &  & Training & Testing & Total \\
\midrule
 & \# Patients & 120 & 50 & 170 \\
 & \# Scans    & 240 & 100 & 340 \\
 & \# Landmarks & 21394 & 9142 & 30536 \\
\midrule
\multirow{3}{*}[-1.69ex]{\makebox[0pt][c]{\rotatebox[origin=c]{90}{Scanners}}} & \# iTero Elem. 2 & 110 & 46 & 156 \\
 & \# Trios3 & 74 & 30 & 104 \\
 & \# Primescan & 56 & 24 & 80 \\
\cmidrule(lr){2-5}
 & \# Total & 240 & 100 & 340 \\
\bottomrule
\end{tabular}
}

\vspace{0.2cm}

\rev{
\begin{tabular}{lcc|lcc}
\toprule
\multicolumn{3}{c|}{\textbf{Upper Jaw}} &
\multicolumn{3}{c}{\textbf{Lower Jaw}} \\
\midrule
Label & Training & Testing & Label & Training & Testing \\
\hline
Cusp    & 2583  & 1108 & Cusp    & 3010 & 1235 \\
Mesial  & 1581  & 680  & Mesial  & 1602 & 694  \\
Distal  & 1577  & 679  & Distal  & 1581 & 687  \\
Inner   & 1578  & 679  & Inner   & 1575 & 672  \\
Outer   & 1572  & 667  & Outer   & 1582 & 683  \\
Facial  & 1573  & 670  & Facial  & 1580 & 688  \\
Total   & 10464 & 4483 & Total   & 10930 & 4659 \\
\bottomrule
\end{tabular}
}
\end{center}

\end{table}

\begin{figure}
    \centering

    \revbox{\includegraphics[width=\linewidth]{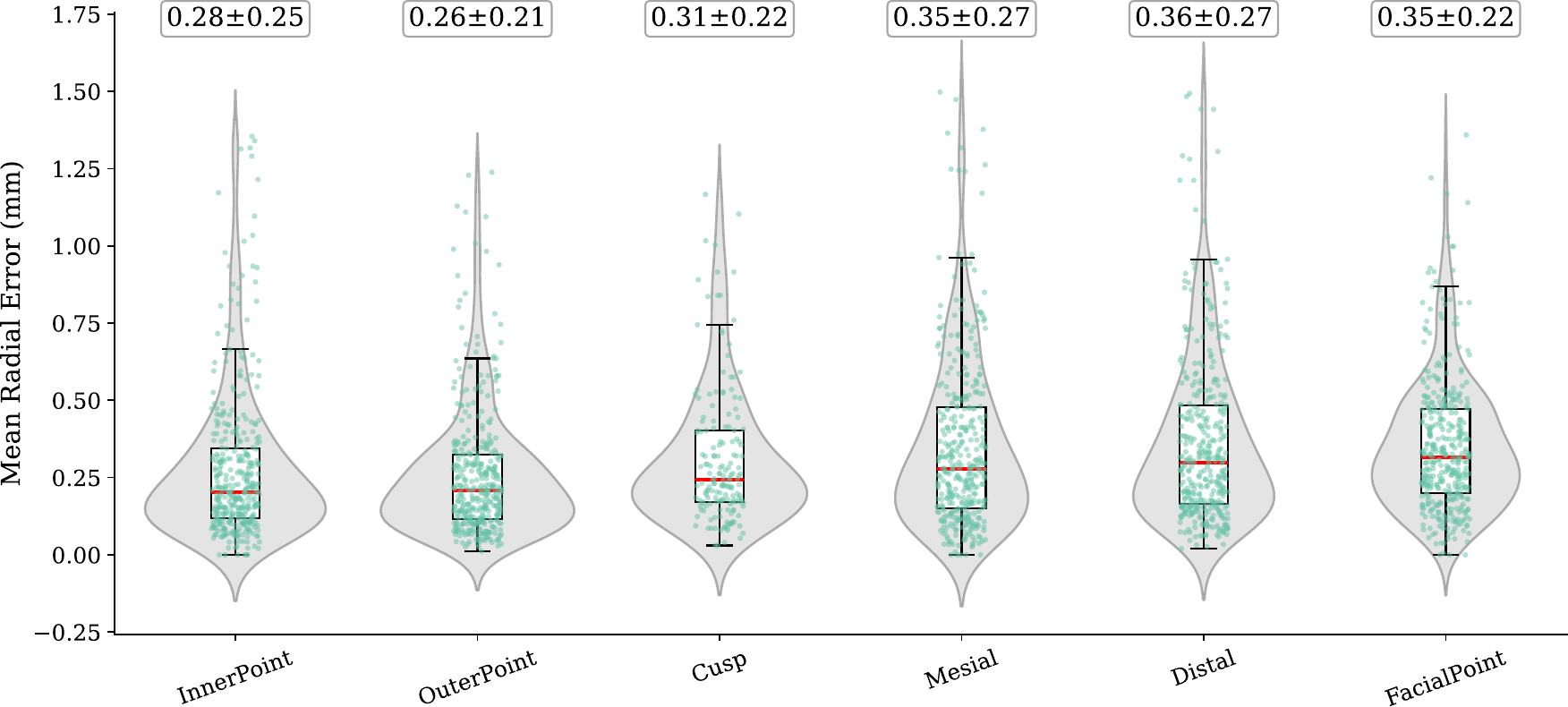}}
    \caption{\rev{Inter-observer variability analysis across different landmark types using Mean Radial Error (MRE).}}
    \label{fig:inter_observer}
\end{figure}

\subsection{\rev{Inter-observer variability analysis}}
\rev{Establishing a human baseline to contextualize the clinical relevance of the reported algorithmic performance is essential. Inter-observer variability was assessed on a representative subset of 40 scans comprising 3,412 annotated landmarks. Rather than random sampling, cases were intentionally selected to capture the diversity of the dataset, including anatomical variability and varying levels of case difficulty. Agreement between the three clinical experts was quantified using the Mean Radial Error (MRE) and the Intra-class Correlation Coefficient (ICC). \figureabvr \ref{fig:inter_observer} presents the obtained MRE for each category. The results indicate an overall human baseline of 0.34 $\pm$ 0.32 mm (median: 0.26 mm), showing that the expected margin of expert annotation variability is approximately 0.3 mm. This provides a clinically meaningful reference for interpreting the performance of the competing methods. In addition, the variability remains consistently low across all landmark categories. The highest precision was observed for outer points (0.26 $\pm$ 0.21 mm), whereas mesial landmarks exhibited the largest relative variability (0.36 $\pm$ 0.27 mm). Landmarks such as inner point and outer point exhibit narrow distributions, reflecting high localization precision, whereas mesial and distal points demonstrate wider distributions, reflecting the increased anatomical complexity and the close proximity of neighboring structures in these regions. Importantly, the challenge evaluation protocol was designed in alignment with this human baseline. The official scoring script evaluates performance across 30 distance thresholds ranging from 0.1 mm to 3.0 mm, ensuring that the reported mAP and mAR metrics capture performance from expert-level precision ($\sim$ 0.3 mm) up to broader clinically acceptable tolerances (3.0 mm). Inter-observer reliability was further evaluated using ICC(2,1), yielding values of 1.000, 0.999, and 0.995 along the x-, y-, and z-axes, respectively, corresponding to a mean ICC of 0.998. These near-perfect ICC values indicate that variability between observers is negligible relative to the anatomical variability across scans, thereby confirming the high reliability and reproducibility of the annotation protocol.}

\subsection{Data records}

Landmark annotations for each intraoral scan in the dataset are delivered in JavaScript Object Notation (JSON) format. \rev{This file (see \figureabvr\ref{fig:record} in the Appendix) provides essential information,} including the version of the file format and the unique identifier for the corresponding intraoral scan. It also contains a list of objects representing the landmark points identified on the scan. Each object is described by a unique key that serves as an identifier for the point, its class, and its 3D coordinates.

\section{Top ranked methods}
\label{sec:methods}
A total of 49 teams officially participated in the 3DTeethLand challenge, with 115 team members preregistered. During the preliminary phase, 10 teams made their submissions, and 6 teams proceeded to the final phase with their final submissions uploaded to the leaderboard. The participating algorithms are required to detect all visible teeth within each intraoral 3D scan. All the algorithms submitted by participants were run on the Synapse platform \footnote{\href{https://www.synapse.org/Synapse:syn57400900/wiki/}{https://www.synapse.org/Synapse:syn57400900/wiki/}}. Submission instructions \footnote{\href{https://www.synapse.org/Synapse:syn57400900/wiki/629180}{https://www.synapse.org/Synapse:syn57400900/wiki/629180}} were published on the webpage to guide participants in submitting their Docker files to the evaluation platform. After submitting their Docker files, the teams received an email indicating whether the submission ran successfully, along with a link where they could view the execution logs and the evaluation result on one randomly selected training case. The training algorithms were not limited to the data provided by the challenge. However, \rev{none of the participants reported using additional external datasets}.

\tableabvr \ref{tab:methods_summary} provides a summary of the top six ranked methods, which will be explored in-depth in the subsequent section for a comprehensive understanding of their approaches. For clarity, a specific color is assigned to each team, which is used consistently in both the tables and the experimental results. These methods can be divided into two main categories. The first category follows the typical setup involving tooth segmentation followed by landmark detection. The second category does not involve tooth segmentation.

\revtable{
\FloatBarrier
\begin{landscape}

\renewcommand{\arraystretch}{2}
\setlength{\tabcolsep}{5pt}
\fontsize{7pt}{8.4pt}\selectfont
\begin{longtable}
{>{\raggedright\arraybackslash}p{2.4cm}>{\raggedright\arraybackslash}p{3cm}>{\raggedright\arraybackslash}p{2.6cm}>{\raggedright\arraybackslash}p{3cm}>{\raggedright\arraybackslash}p{3.3cm}>{\raggedright\arraybackslash}p{5cm}}
\caption{Summary of the participating methods}
\label{tab:methods_summary} \\
\toprule
\multicolumn{1}{l}{Team/Ref. Authors} & 
\multicolumn{1}{l}{Preproc.} & 
\multicolumn{1}{l}{ Data aug.} & 
\multicolumn{1}{l}{Postproc.} & 
\multicolumn{1}{l}{Architectures} & 
\multicolumn{1}{l}{Loss functions} \\
\midrule
\endfirsthead
\toprule
\multicolumn{6}{c}{\textit{Table \ref{tab:methods_summary} (continued)}}\\
\midrule
\multicolumn{1}{c}{Team/Ref. Authors} & 
\multicolumn{1}{c}{Preproc.} & 
\multicolumn{1}{c}{ Data aug.} & 
\multicolumn{1}{c}{Postproc.} & 
\multicolumn{1}{c}{Architectures} & 
\multicolumn{1}{c}{Loss functions} \\
\midrule
\endhead
\midrule

\endfoot
\bottomrule
\endlastfoot
\fcolorbox{white}{lightpink}{\rule{0pt}{5pt}\rule{5pt}{0pt}}\quad 
Radboud (Niels an Nistelrooij \etal) &
Z-score normalization, pose normalization &
Random horizontal flipping, scaling, and rotation around the z-axis &
Single-tooth projection; weighted DBSCAN; surface refinement &
\parbox[t]{\linewidth}{\raggedright \tabitem Tooth Seg.: Stratified Transformer\par \tabitem Landmark Det.: Stratified Transformer} &
\tabitem Tooth Seg.: spatial embedding loss; cross-entropy (CE); focal loss
\newline
\tabitem Landmark Det.: binary CE loss for the binary segmentation; smooth-L1 loss + Chamfer distance loss + separation loss for the landmarks\\
\fcolorbox{white}{olive}{\rule{0pt}{5pt}\rule{5pt}{0pt}}\quad 
YY-LAB (Kaibo Shi \etal) &
Mesh simplification ($\sim$10k facets); landmark heatmap computation &\multicolumn{1}{c}{\xmark } 
 &
\multicolumn{1}{c}{\xmark } &
\parbox[t]{\linewidth}{\raggedright \tabitem Tooth Seg.: TeethGNN \newline 
\tabitem Landmark Det.: TeethGNN encoder + dual decoders (fixed and cusp) }&
\tabitem Tooth Seg.: N/A\newline \tabitem Landmark Det.: bipartite matching loss (CE for probabilities loss and mean squared error (MSE) for heatmap loss) \\
\fcolorbox{white}{pink}{\rule{0pt}{5pt}\rule{5pt}{0pt}}\quad  YN-LAB (Zhu Xiaoying \etal)  & Gingiva removal; FPS downsampling to 30,000 pts; normalization to unit sphere  & Vertex jittering & HDBSCAN clustering per landmark category; Gaussian weighted voting & \parbox[t]{\linewidth}{\raggedright \tabitem Tooth Seg.: 3D UNET  \newline 
\tabitem Landmark Det.:Multi-stage PointMLP}  & \tabitem Landmark Det.: Dice Score Coefficient (DSC) \\
\fcolorbox{white}{lightgreen}{\rule{0pt}{5pt}\rule{5pt}{0pt}}\quad     IGIP-LAB (Weijie Liu \etal)  & Point Sampling; Feature Extraction; Distance Field Computation; Confidence Mapping; confidence level computation &  Grid sampling; random offset perturbations&  Confidence Filtering; Density-based Clustering;Final Landmark Selection & \parbox[t]{\linewidth}{\raggedright \tabitem Tooth Seg.: N/A\newline \tabitem Landmark Det.: PointTransformer V3} & \parbox[t]{\linewidth}{\raggedright  \tabitem Tooth Seg.: N/A\newline \tabitem Landmark Det.: MSE }\\
 \fcolorbox{white}{lightblue}{\rule{0pt}{5pt}\rule{5pt}{0pt}}\quad  ChohoTech (Huikai Wu) &  FPS Resampling to 20,000 points; Feature Extraction (patch centers, normal vectors, and vertex coordinates) &Random perturbations & Threshold Filtering; Non-Maximum Suppression & \parbox[t]{\linewidth}{\tabitem Tooth Seg.: N/A \newline \tabitem Landmark Det.: DGCNN} & \parbox[t]{\linewidth}{\raggedright \tabitem Tooth Seg.: N/A\newline  \tabitem Landmark Det.: L2 loss; L1 loss}\\
  \fcolorbox{white}{rose}{\rule{0pt}{5pt}\rule{5pt}{0pt}}\quad   3DIMLAND (Tibor Kubík \etal)  &  Randomly sampling 64k mesh vertices; enriching vertices with normal information; converting landmark positions into per-vertex geodesic distance maps & Random rotations, translations, isotropic scaling, and Free-Form Deformation (FFD)–based geometric morphing & Calibrated topology-driven non-minima suppression; topology-aware graph updates; distance thresholding for final landmark extraction & \parbox[t]{\linewidth}{\raggedright \tabitem Tooth Seg.: N/A\newline  \tabitem Landmark Det.: Point Transformer v3–based encoder–decoder} &   \parbox[t]{\linewidth}{\raggedright \tabitem Tooth Seg.: N/A\newline  \tabitem Landmark Det.: MSE}\\
\end{longtable}
\normalsize
\end{landscape}
\FloatBarrier
}

\begin{figure}[h]
    \centering
    \revbox{
    \includegraphics[width=0.8\linewidth]{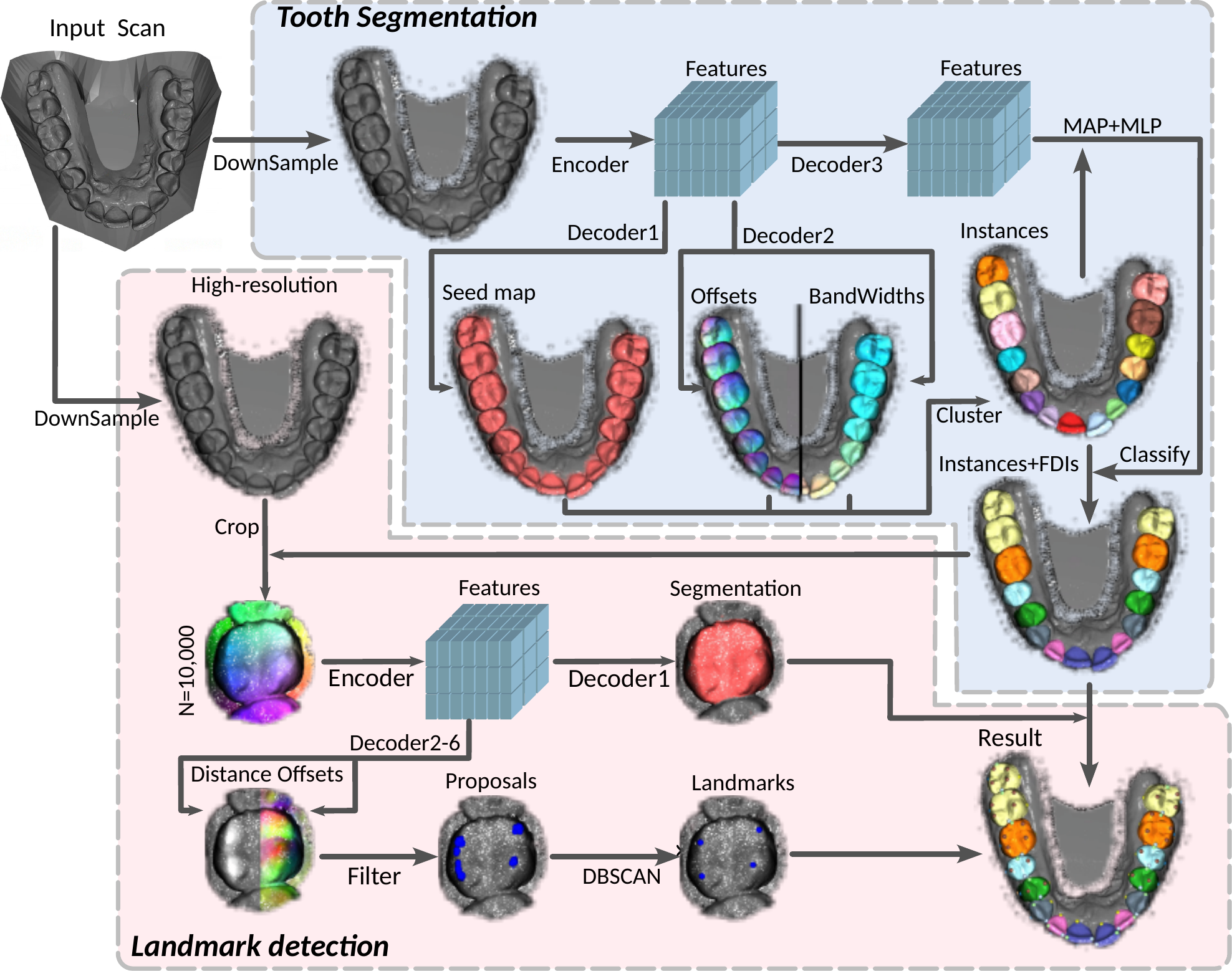}}
    \caption{\rev{Overview of the method proposed by the Radboud team. The framework consists of two stages: (1) tooth instance segmentation and labeling using a shared encoder with multiple decoders operating on a downsampled point cloud to estimate seed maps, offsets, and instance embeddings. (2) Tooth landmark detection performed on high-resolution tooth crops using dedicated decoders to predict different landmark classes.}}
    \label{fig:niels_et_al}
\end{figure}

\subsection{\fcolorbox{white}{lightpink}{\rule{0pt}{9pt}\rule{9pt}{0pt}} Radboud team (Niels van Nistelrooij \etal)}

The proposed approach by Niels van Nistelrooij \etal \cite{van2024toothinstancenet} is shown in \figureabvr \ref{fig:niels_et_al}. Their method follows a two-stage process. The first stage performs end-to-end tooth instance segmentation and labeling, leveraging a large context specifically optimized for accurate tooth labeling. In the second stage, high-resolution tooth crops are analyzed to refine the tooth segmentation and to independently predict landmarks of each landmark class.

\subsubsection{Tooth instance segmentation}
In this step, the surface mesh is first transformed into a downsampled point cloud to reduce the number of points for processing. Then, a shared encoder along with three separate decoders are employed. The first decoder generates a seed map for binary semantic segmentation of teeth, with higher seed values indicating points closer to a tooth center. The second decoder estimates offsets from points to their nearest tooth center as well as bandwidths. Tooth instances are identified iteratively, selecting the highest seed value to define a Gaussian distribution representing the tooth's position and size. Points with a probability above 0.5 based on this Gaussian distribution are clustered to form a tooth instance. This process continues until no high seed values remain. The third decoder extracts features from the encoder for tooth labeling, averaging features of each instance to create an instance embedding by masked average pooling, which is processed by a multi-layer perceptron to predict tooth labels. Finally, by combining tooth instance segmentation and tooth labeling, a multiclass tooth instance segmentation is obtained.

\subsubsection{Tooth landmark detection} 
The model dedicated to landmark detection processes single-tooth crops and is made up of one shared encoder and six separate decoders. The first decoder generates a high-resolution binary segmentation of the center tooth, enhancing the accuracy of the low-resolution segmentation produced during the tooth instance segmentation stage. Decoders two through six were tasked with detecting specific landmarks: mesial and distal (decoder two), facial (decoder three), inner (decoder four), outer (decoder five), and cusp landmarks (decoder six). By assigning different decoders to distinct landmark classes, the model achieved improved accuracy in positioning each landmark. For example, decoder six identifies the nearest cusp landmark by estimating distances and offsets for each point. Points within a defined distance threshold are adjusted based on their offsets to create landmark proposals. These proposals are then clustered using the weighted DBSCAN algorithm \cite{ester1996density}, where the weights are inversely related to the predicted distances. A landmark's final position is determined by calculating the weighted average of the cluster's landmark proposals. Lastly, mesial and distal landmarks from decoder two are categorized using a rule-based approach, relying on the tooth's FDI number and the landmarks' relative position to the tooth center.

\subsection{\fcolorbox{white}{olive}{\rule{0pt}{9pt}\rule{9pt}{0pt}} YY-LAB team (Kaibo Shi \etal)}
\begin{figure}[t]
    \centering
    \revbox{
    \includegraphics[width=0.95\linewidth]{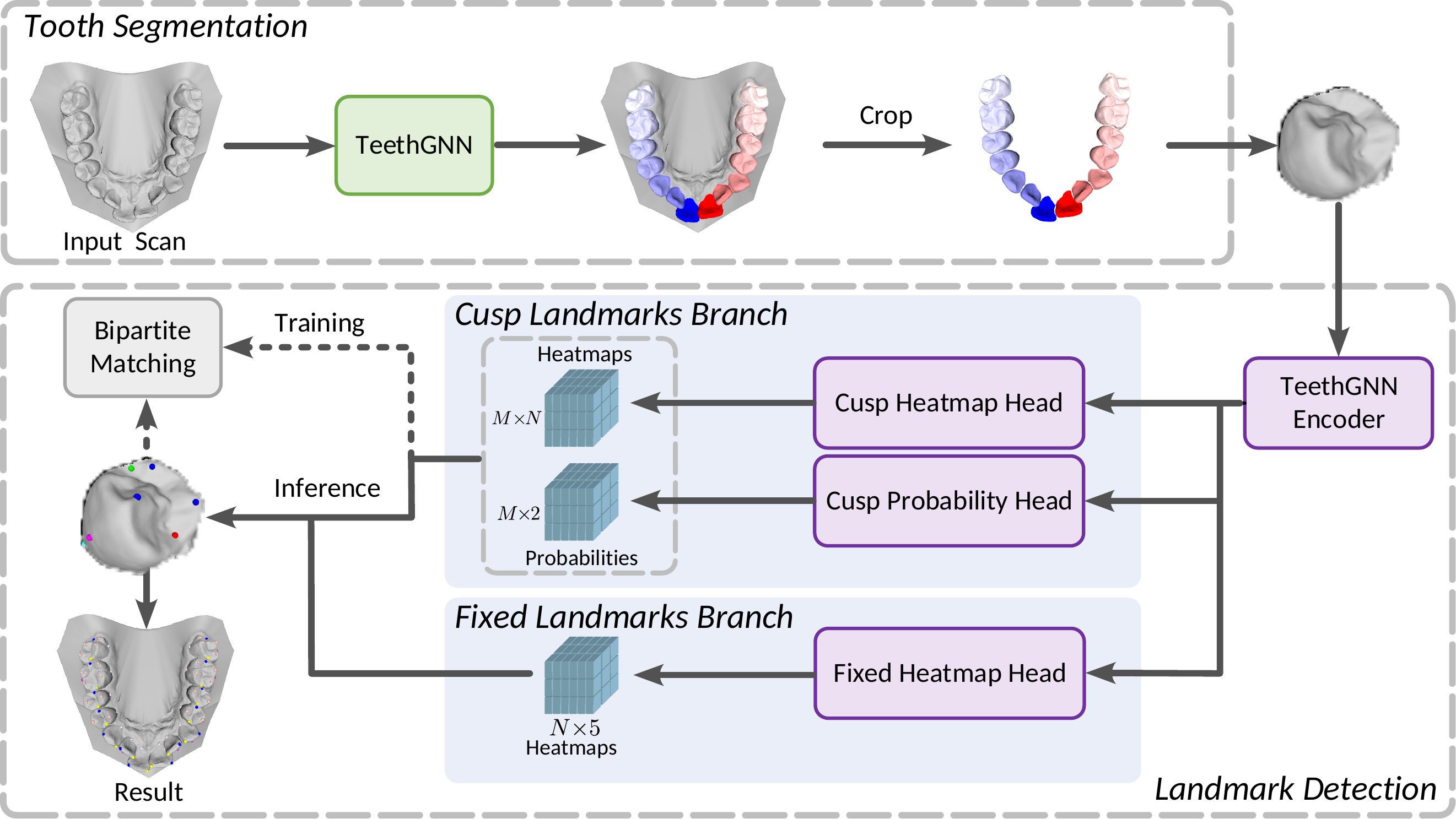}}
    \caption{\rev{Overview of the method proposed by the YY-LAB team. The framework follows a two-stage pipeline: (1) 3D tooth segmentation on decimated jaw meshes using the TeethGNN architecture with graph-cut refinement to obtain individual tooth patches. (2) Landmark detection using the TL-DETR network, which predicts fixed landmarks and variable cusp landmarks through specialized decoder branches.}}
    \label{fig:yylab}
\end{figure}

Since landmark detection directly from a jaw model is highly challenging, the YY-LAB team proposed 3D tooth segmentation as an initial phase. As shown in \figureabvr \ref{fig:yylab}, the proposed method by Kaibo Shi \etal consists of two stages: 3D tooth segmentation and landmark detection.

\subsubsection{Tooth Segmentation}
To address challenges during network training and inference, the authors proposed reducing the resolution of jaw meshes from over 100,000 facets to approximately 10,000 using mesh decimation
algorithm \cite{garland1997surface}. Features extracted from the facets, comprising centroid coordinates, facet normals, and vectors from each of the three vertices to the centroid, will be the input into the TeethGNN model \cite{zheng2022teethgnn}. This model generates semantic labels, which are further refined using the graph-cut algorithm to improve boundary definition. After projecting back the refined labels onto the original mesh, the teeth will be cropped into tooth patches from the jaw model according to these segmentation results.

\begin{figure}[t]
    \centering
    \revbox{
    \includegraphics[width=\linewidth]{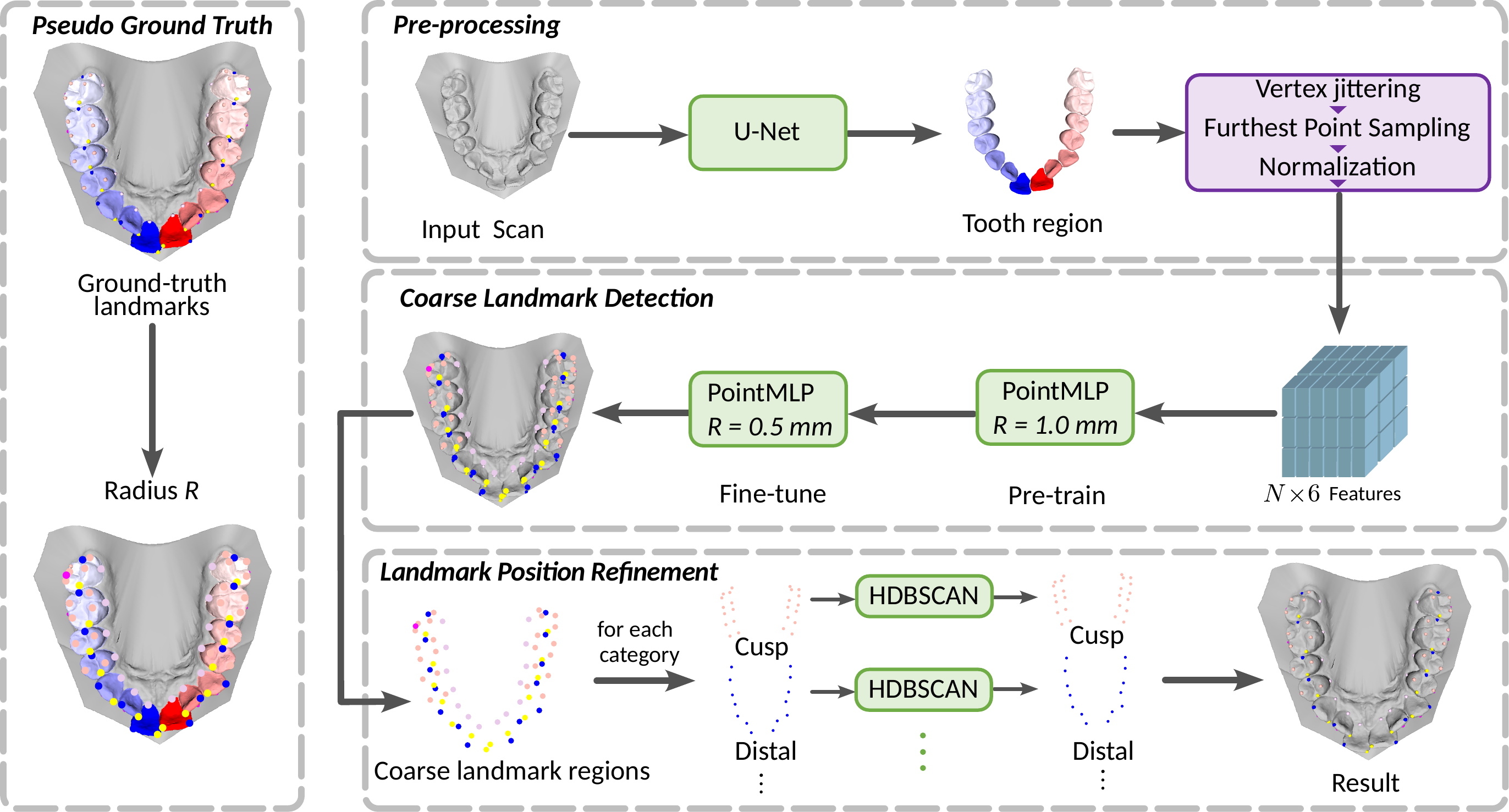}}
    \caption{\rev{Overview of the YN-LAB pipeline. The method employs curriculum learning and consists of two stages: (1) tooth-region extraction and point-cloud preprocessing. (2) Coarse-to-fine landmark detection using PointMLP segmentation, followed by HDBSCAN clustering and Gaussian weighted voting for final landmark localization.}
    }
    \label{fig:ynlab}
\end{figure}

\subsubsection{Landmark detection}
For landmark detection for individual teeth, the authors were inspired by the KeypointDETR~\cite{jin2025keypointdetr} and proposed TL-DETR. In this network, two decoder branches are considered: a fixed landmark branch and a cusp landmark branch, as the number of cusp points varies based on the tooth's shape and type, while other landmarks are fixed on every tooth. Given a single tooth patch, TL-DETR predicts heatmaps for the fixed landmarks and both heatmaps and probabilities for the potential cusp landmarks. Following a process similar to tooth segmentation, facet features are first extracted and fed into the network. These features are processed by the TeethGNN encoder, which utilizes static EdgeConv layers to capture local geometric information. Fully connected and pooling layers then generate global features, which are combined with the local features and decoded through another EdgeConv layer to produce the final feature representation. The decoding process consists of two branches tailored for fixed and cusp landmarks. For fixed landmarks, the feature representation is passed to a heatmap head comprising an MLP and Conv1d layers, producing heatmaps for the five fixed landmarks. For cusp landmarks, the feature is decoded into a heatmap feature, which is then processed by separate heatmap and probability heads. These heads generate heatmaps representing cusp landmark locations and binary probabilities distinguishing between cusp landmarks and background points. The bipartite matching loss is used to measure the discrepancy between the prediction sets and the ground truth cusp heatmaps, facilitating end-to-end inference.

\begin{figure}[t]
    \centering
    \revbox{
    \includegraphics[width=\linewidth]{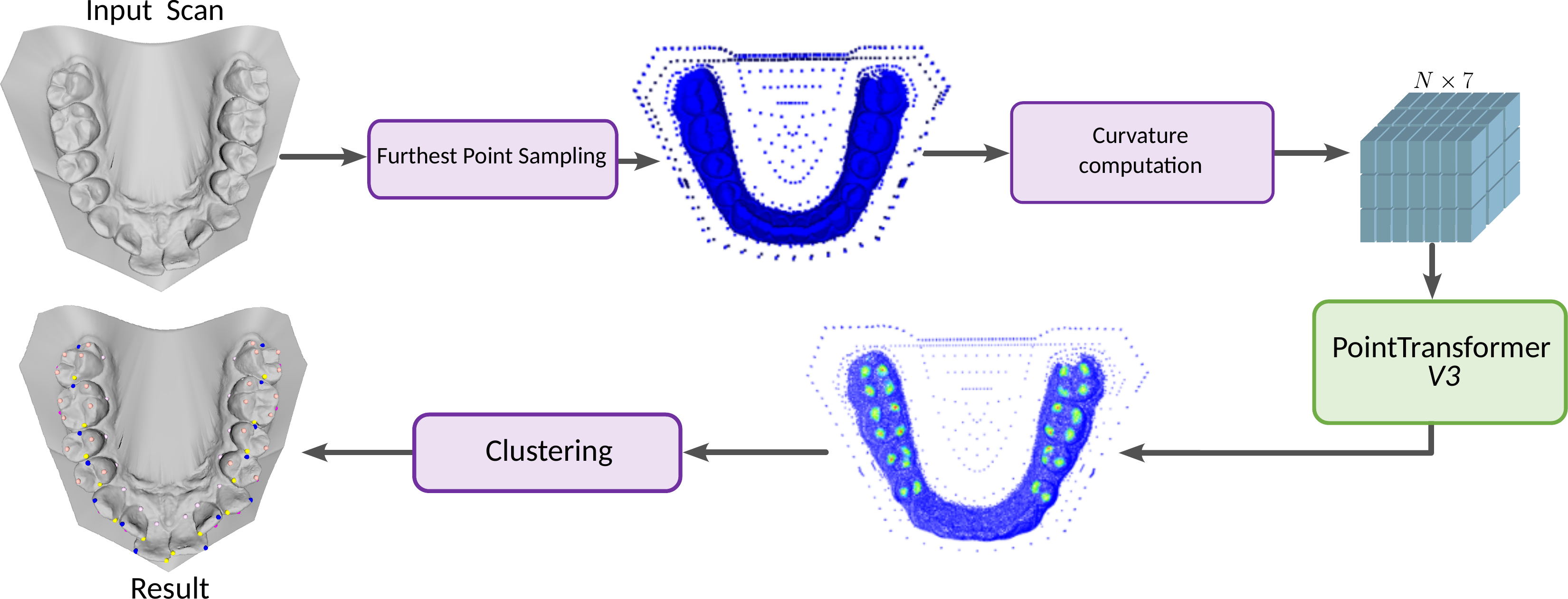}}
    \caption{\rev{Overview of the IGIP-LAB pipeline. The method operates directly on 3D tooth models converted into point clouds and consists of three main stages: (1) Point-cloud preprocessing with FPS resampling and feature extraction. (2) Landmark prediction using PointTransformer V3. (3) Post-processing with confidence thresholding followed by density-based clustering for final landmark localization.}}
    \label{fig:IGIP}
\end{figure}

\subsection{\fcolorbox{white}{pink}{\rule{0pt}{9pt}\rule{9pt}{0pt}}  YN-LAB team (Zhu Xiaoying \etal)}
 The global pipeline of the YN-LAB team is presented in \figureabvr \ref{fig:ynlab}. The proposed method adopts the curriculum learning paradigm \cite{bengio2009curriculum}, in which the training process is organized to gradually tackle tasks of increasing complexity. It begins with an initial segmentation task targeting larger regions of interest around the ground-truth landmarks, and it progresses to a more detailed segmentation task focusing on smaller regions.

\subsubsection{Data Pre-processing}
In this step, an input 3D jaw model was processed to segment out the tooth region only, which addresses potential data imbalance since the tooth region occupies a small region over the entire surface of the jaw model. Moreover, all tooth landmarks are located within the tooth region. This tooth segmentation process was done using the coarse-to-fine segmentation algorithm proposed by the TeethSeg in the 3DTeethSeg'22 challenge \cite{ben20233dteethseg, dascalu2023assignment}, which refines the tooth-gingival boundary to achieve accurate tooth segmentation. To reduce model overfitting, the segmented tooth region undergoes several steps of data augmentation. First, vertex jittering was applied by adding Gaussian noise at several levels of intensity to the points in the segmented regions, which potentially covers tooth shape variations. Next, Farthest Point Sampling (FPS) \cite{li2022adjustable} was used to ensure a well-distributed distribution of downsampled points from the segmented tooth region. Lastly, the downsampled points were normalized to be within a unit sphere, and their normal vectors were computed.

\subsubsection{Landmark detection}
This step is composed of two main phases: coarse landmark detection and precise landmark detection. Coarse landmark detection aims to segment the region around the landmark. It takes $N$ normalized points, which include both points' coordinates and normal vectors. Then, it leverages the PointMLP network \cite{ma2022rethinking} to effectively capture both the point cloud's local and global features and perform the segmentation task. The training process is guided by pseudo ground truths, i.e., points within radius $R$ from each landmark were assigned the landmark's label. Empirically, the authors found that the network was unable to perform accurate segmentation with small $R$ directly, potentially due to severe data imbalance. Therefore, they adopted a curriculum learning approach to enhance the segmentation accuracy. Initially, the network was pre-trained on a pseudo ground truth with a radius $R$ of 1.0 mm until convergence. Next, the pre-trained network was fine-tuned on pseudo ground truth with a smaller radius $R$ of 0.5 mm until convergence, which is a more complex task. This curriculum learning-inspired approach gives more accurate segmentation results.

The precise landmark detection stage uses the segmented region from the coarse landmark detection to determine the precise location of each landmark within it. In this step, the HDBSCAN clustering algorithm \cite{tran2021hdbscan} was employed to group segmented points belonging to a specific landmark's category based on their points' density. After the clustering, Gaussian weighted voting is used. This method assigns higher importance to points closer to the cluster center, which serves as the detected tooth landmark.

\subsection{\fcolorbox{white}{lightgreen}{\rule{0pt}{9pt}\rule{9pt}{0pt}} IGIP-LAB team (Weijie Liu \etal)}
The global pipeline of the proposed method by Weijie Liu \etal, which does not include a segmentation step, is presented in \figureabvr \ref{fig:IGIP}. In their approach, the authors proposed transforming 3D tooth models into point cloud data to enhance the model's ability to capture complex geometries. Therefore, they started by sampling 16384 points from each model to obtain the corresponding point cloud data, which includes both coordinate and normal information using the FPS. Next, the curvature of each point is calculated using the obtained normal data. In addition, the Euclidean distance field between landmarks and corresponding teeth is calculated. This distance field is then used to guide the network training by translating these distances into confidence levels. With the PointTransformer V3 as the foundational architecture, the network will be used to predict the confidence levels for various landmark categories. The final step is a post-processing step that filters out landmarks with confidence values below 0.7. The remaining landmarks are grouped via density-based clustering, with the highest-confidence landmark in each cluster selected as the final landmark. 

\begin{figure}[t]
    \centering
    \revbox{
    \includegraphics[width=\linewidth]{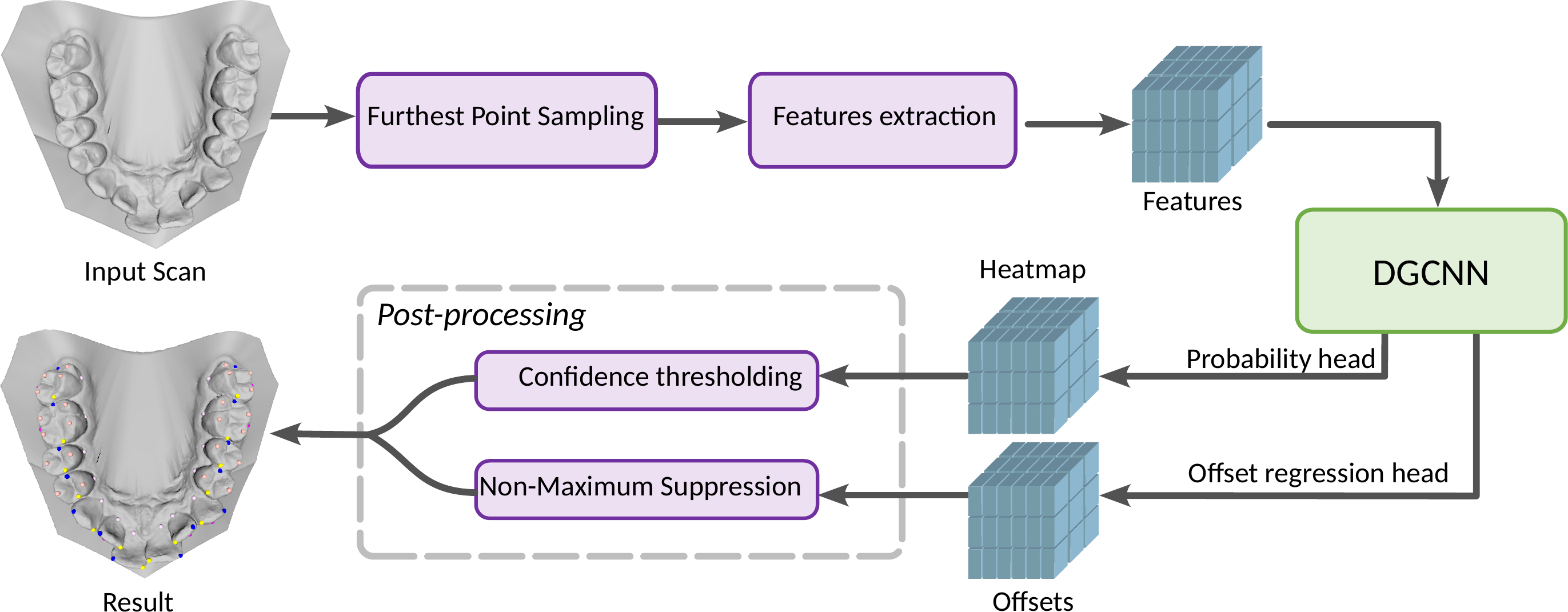}}
    \caption{\rev{Overview of the ChohoTech pipeline. The method operates directly on 3D intraoral scans converted into point clouds and consists of three main stages: (1) Point-cloud preprocessing with FPS resampling and feature extraction. (2) Landmark prediction using a DGCNN with dual output branches for probability heatmaps and offset regression. (3) Post-processing with confidence thresholding and Non-Maximum Suppression for final landmark localization.}}
    \label{fig:chochotech}
\end{figure}

\subsection{ \fcolorbox{white}{lightblue}{\rule{0pt}{9pt}\rule{9pt}{0pt}} ChohoTech team (Huikai Wu)}
The proposed approach by Huikai Wu also does not involve segmentation. An overview of his approach is depicted in \figureabvr \ref{fig:chochotech}. This 3D dental landmark detection method involves several key steps, beginning with data preprocessing. First, 3D intraoral scan models are resampled using FPS to generate consistent point clouds, with each patch characterized by its center coordinates, normal vector, and vertices. The following features will serve as input to the proposed network. Its architecture is based on Dynamic Graph Convolutional Networks (DGCNN), which includes two output branches: one for estimating the likelihood of a point being a landmark (probability heatmap) and another for calculating the residual vector from each point to its nearest landmark (offset regression). At each layer, the network performs a K-Nearest Neighbor (KNN) search based only on the input point coordinates, enabling it to capture local geometric relationships. Finally, a post-processing stage is performed that applies confidence thresholding to filter out low-confidence points. In addition, a Non-Maximum Suppression (NMS) is applied to eliminate redundant detections, ensuring precise landmark identification.

\subsection{ \fcolorbox{white}{rose}{\rule{0pt}{9pt}\rule{9pt}{0pt}} 3DIMLAND team (Tibor Kubík \etal) } 

\begin{figure}[t]
    \centering
    \revbox{
    \includegraphics[width=\linewidth]{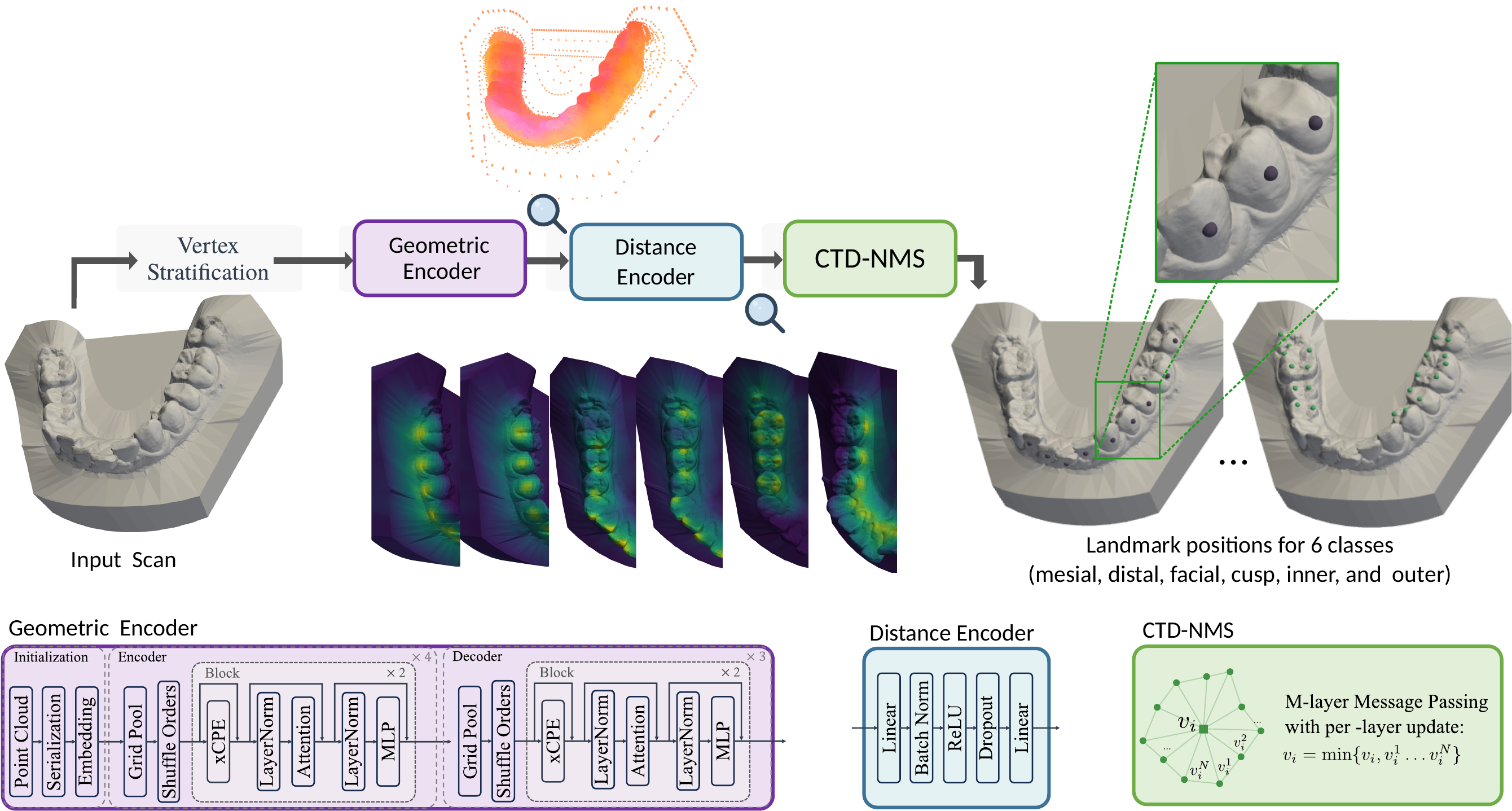}}
    \caption{\rev{Overview of the 3DIMLAND pipeline. The method predicts dental landmarks directly from 3D meshes and consists of three stages: (1) Geometry encoding with point sampling and vertex normal augmentation. (2) Distance decoding to produce six-channel per-point distance maps for multiple landmark categories. (3) Graph-based Calibrated Topology-Driven Non-Minima Suppression (CTD-NMS) for refined landmark extraction and false positive removal.}}
    \label{fig:3DIMLAND}
\end{figure}
\figureabvr \ref{fig:3DIMLAND} depicts the global pipeline of the method proposed by Tibor Kubík \etal \cite{kubik2024leveraging}. This method leverages transformer architectures to model a function that predicts landmark positions from an input mesh representing a single 3D dental scan. The framework is composed of three key components: a geometry encoder, a distance decoder, and a graph-based non-minima suppression module. The role of the Geometry Encoder step is to process the dental mesh into a meaningful, latent feature space. Since dental meshes have varying numbers of vertices, this step ensures consistency by randomly selecting a fixed number of points from the original mesh, reducing the complexity caused by varying mesh resolutions. To avoid sampling irrelevant mesh regions, the proposed method ensures that the sampled points are focused primarily on the important parts of the mesh, such as the teeth and gingiva, rather than unnecessary structures. Additionally, it enriches the point cloud by concatenating vertex normal information, providing the model with more geometric context. The Distance Decoder in this framework takes the learned per-point features from the Geometry Encoder and maps them into a 6-channel per-point distance output. This design allows the framework to detect landmarks in six different categories without needing intra-class label assignment. The Calibrated Topology-Driven Non-Minima Suppression (CTD-NMS) module is employed to refine the landmark extraction process from the distance decoder's output. While a basic approach might involve extracting the maxima from the thresholded distance maps, it is prone to producing clusters of false positives, particularly when the input geometry provides insufficient information. To address this issue, the authors proposed the CTD-NMS, which uses a graph-convolutional operator to iteratively update each vertex's value by considering its neighbors in the mesh, ensuring only the most robust landmarks are retained. A vertex is classified as a landmark if its value remains unchanged and its distance is below a predefined threshold, with the parameters calibrated based on model performance and mesh resolution.

\section{Experimental results}
\label{sec:results}
In this section, we detail the employed evaluation metrics, describe the ranking protocol, and present the obtained results.

\subsection{Evaluation metrics}
Within the existing literature, various metrics such as root mean square error (RMSE) \cite{triarjo2023automatic}, mean distance error (MDE) \cite{wei2022dense}, and mean absolute error (MAE) \cite{wu2022two}, have been commonly used for the assessment of landmark localization,  typically assuming a fixed number of landmarks. However, our challenge context differs from this norm as we are dealing with a detection task involving a variable number of landmarks per tooth. In this context, and in line with \cite{you2020keypointnet} and \cite{maier2024metrics}, we considered two metrics, the Mean Average Precision (mAP) and the Mean Average Recall (mAR), as these metrics are more suitable and align more effectively with the nature of this task. To evaluate the accuracy of landmark detection, we categorize the landmarks into four distinct groups according to their clinical relevance, as outlined below:

\begin{multicols}{2}

\begin{itemize}
    \item Category 1: Mesial/Distal
    \item Category 2: Cusps
    \item Category 3: Outer/Inner
    \item Category 4: Facial

\end{itemize}
\end{multicols}
The mAP is computed by aggregating the Average Precision (AP) values across different landmark categories. Each AP is derived from the area under the Precision-Recall curve and evaluated at multiple distance thresholds. As a result, we calculate a distinct mAP value for each landmark category, providing a comprehensive assessment of detection performance across the different landmarks.

Similarly, the mAR is calculated by aggregating the Average Recall (AR) for each category, where AR is determined by the area under the Recall-exp(-Distance) curve. As with mAP, the number of mAR values corresponds to the number of landmark categories.

The localization criterion is defined by the Euclidean distance between predicted and reference landmarks. A prediction is considered a valid detection (or ``hit'') if the Euclidean distance between the predicted and reference landmarks falls below a predefined threshold. Given the varying clinical requirements for landmark localization, multiple thresholds are used to assess the performance of the detection algorithm. These thresholds range from 0 mm to 3 mm, incremented by 0.1 mm, capturing a broad spectrum of practical scenarios.

For the landmark assignment, a greedy strategy is employed, wherein predicted landmarks are ranked based on their predicted class scores. The assignment proceeds by iteratively matching each predicted landmark, starting from the highest score to the nearest reference landmark. It is important to note that each reference landmark can only be assigned to one predicted landmark, ensuring a one-to-one correspondence. This strategy optimizes the matching process while maintaining robustness in the evaluation of detection accuracy.

\subsection{Ranking protocol}

To ensure robust rankings, we employed a point-based ranking method enhanced by bootstrapping. The process begins with the computation of the mAP and mAR metrics for each landmark category. Teams are then pairwise compared for each metric using the Wilcoxon Signed Rank Test. A team is awarded one point for each comparison where it is deemed statistically superior (\textit{p}-value $< 0.001$), resulting in a ``total point count" that reflects the number of comparisons won. Bootstrapping is applied by resampling 10\% of the data and repeating the pairwise comparison process on the remaining data, generating a ``total point count" for each resampling iteration. This process is repeated 100 times. The final point score for each team is normalized by the total number of comparisons, calculated as the normalized scores are then aggregated to produce the final ranking, ensuring a statistically robust and fair evaluation of performance.

\begin{table}[b]
\renewcommand{\arraystretch}{1.6}
\centering
\resizebox{\textwidth}{!}{
\rev{
\begin{tabular}{@{}lccccccccccc@{}}
\toprule
\multirow{2}{*}{\textbf{Team}} & \multicolumn{4}{c}{\textbf{AP}} & \multicolumn{4}{c}{\textbf{AR}} & \multirow{2}{*}{\textbf{mAP}} & \multirow{2}{*}{\textbf{mAR}} & \multirow{2}{*}{\textbf{RS}} \\
\cmidrule(lr){2-5} \cmidrule(lr){6-9}
 & \textbf{C} & \textbf{F} & \textbf{I/O} & \textbf{M/D} & \textbf{C} & \textbf{F} & \textbf{I/O} & \textbf{M/D} &  &  &  \\
\midrule
\fcolorbox{white}{lightpink}{\rule{0pt}{5pt}\rule{5pt}{0pt}} 
Radboud
& \textbf{0.77 $\pm$ 0.06} & \textbf{0.76 $\pm$ 0.05} & \textbf{0.79 $\pm$ 0.05} & \textbf{0.79 $\pm$ 0.04}
& \textbf{0.67 $\pm$ 0.03} & \textbf{0.63 $\pm$ 0.04} & \textbf{0.66 $\pm$ 0.05} & \underline{0.65 $\pm$ 0.04}
& \textbf{0.78 $\pm$ 0.04} & \textbf{0.65 $\pm$ 0.04} & \textbf{0.91} \\
\midrule
\fcolorbox{white}{lightblue}{\rule{0pt}{5pt}\rule{5pt}{0pt}} 
ChohoTech
& \underline{0.76 $\pm$ 0.07} & \underline{0.76 $\pm$ 0.07} & \underline{0.78 $\pm$ 0.06} & \underline{0.78 $\pm$ 0.06}
& \underline{0.62 $\pm$ 0.06} & \underline{0.58 $\pm$ 0.08} & \underline{0.62 $\pm$ 0.06} & \textbf{0.67 $\pm$ 0.08}
& \underline{0.77 $\pm$ 0.07} & \underline{0.63 $\pm$ 0.07} & \underline{0.83} \\
\midrule
\fcolorbox{white}{olive}{\rule{0pt}{5pt}\rule{5pt}{0pt}}  
YY-LAB
& 0.68 $\pm$ 0.10 & 0.72 $\pm$ 0.08 & 0.74 $\pm$ 0.07 & 0.70 $\pm$ 0.08
& 0.55$\pm$ 0.09 & 0.56 $\pm$ 0.08 & 0.60 $\pm$ 0.07 & 0.57 $\pm$ 0.07
& 0.71 $\pm$ 0.07 & 0.57 $\pm$ 0.07 & 0.62 \\
\midrule 
\fcolorbox{white}{pink}{\rule{0pt}{5pt}\rule{5pt}{0pt}} 
YN-LAB
& 0.75 $\pm$ 0.12 & 0.66 $\pm$ 0.10 & 0.61 $\pm$ 0.14 & 0.65 $\pm$ 0.10
& 0.53 $\pm$ 0.10 & 0.52 $\pm$ 0.09 & 0.51 $\pm$ 0.11 & 0.53 $\pm$ 0.08
& 0.64 $\pm$ 0.10 & 0.52 $\pm$ 0.08 & 0.31 \\
\midrule
\fcolorbox{white}{lightgreen}{\rule{0pt}{5pt}\rule{5pt}{0pt}} 
IGIP-LAB
& 0.63 $\pm$ 0.10 & 0.59 $\pm$ 0.11 & 0.63 $\pm$ 0.11 & 0.52 $\pm$ 0.16
& 0.51 $\pm$ 0.09 & 0.44 $\pm$ 0.09 & 0.50 $\pm$ 0.09 & 0.41 $\pm$ 0.13
& 0.59 $\pm$ 0.10 & 0.46 $\pm$ 0.08 & 0.13 \\
\midrule
\fcolorbox{white}{rose}{\rule{0pt}{5pt}\rule{5pt}{0pt}}
3DIMLAND
& 0.59 $\pm$ 0.08 & 0.62 $\pm$ 0.06 & 0.55 $\pm$ 0.09 & 0.57 $\pm$ 0.07
& 0.45 $\pm$ 0.06 & 0.45 $\pm$ 0.06 & 0.45 $\pm$ 0.07 & 0.45 $\pm$ 0.05
& 0.57 $\pm$ 0.06 & 0.43 $\pm$ 0.05 & 0.03 \\
\bottomrule
\end{tabular}
}
}
\caption{\rev{Comparison of team performances in terms of Average Precision (AP) and Average Recall (AR) across four anatomical categories: cusps (C), facial (F), inner/outer (I/O), and mesial/distal (M/D). The mean Average Precision (mAP) and mean Average Recall (mAR) are also reported. RS denotes the overall ranking score.}}
\label{tab:ar_ap_rank}
\end{table}

\begin{figure*}[t]
    \centering
    \begin{adjustbox}{max width=0.8\textwidth} 
        \begin{minipage}{\textwidth}
            \begin{subfigure}[t]{0.49\textwidth}
                \centering
                \includegraphics[width=\textwidth]{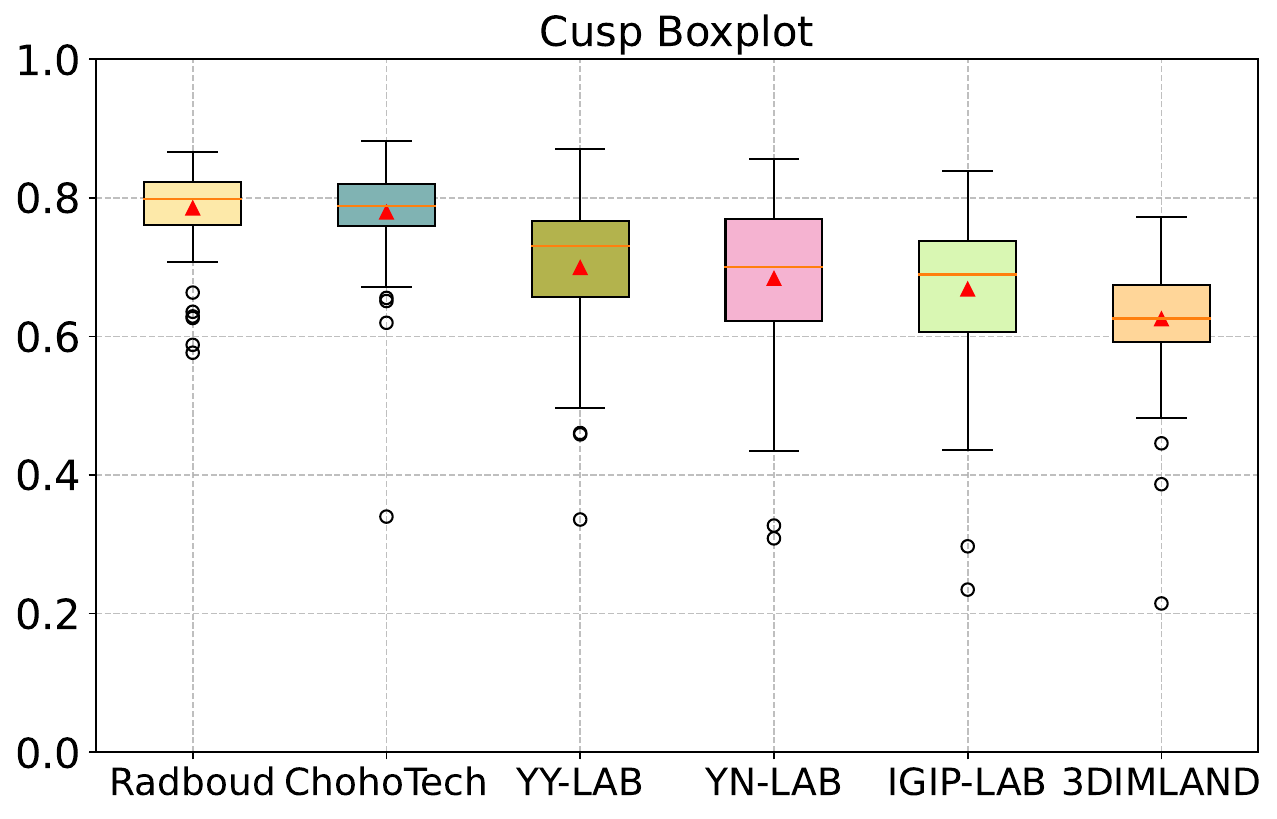}
                \caption{mAP for cusp category}
                \label{fig:cusp_boxplot_mAP}
            \end{subfigure}\hfill
            \begin{subfigure}[t]{0.49\textwidth}
                \centering
                \includegraphics[width=\textwidth]{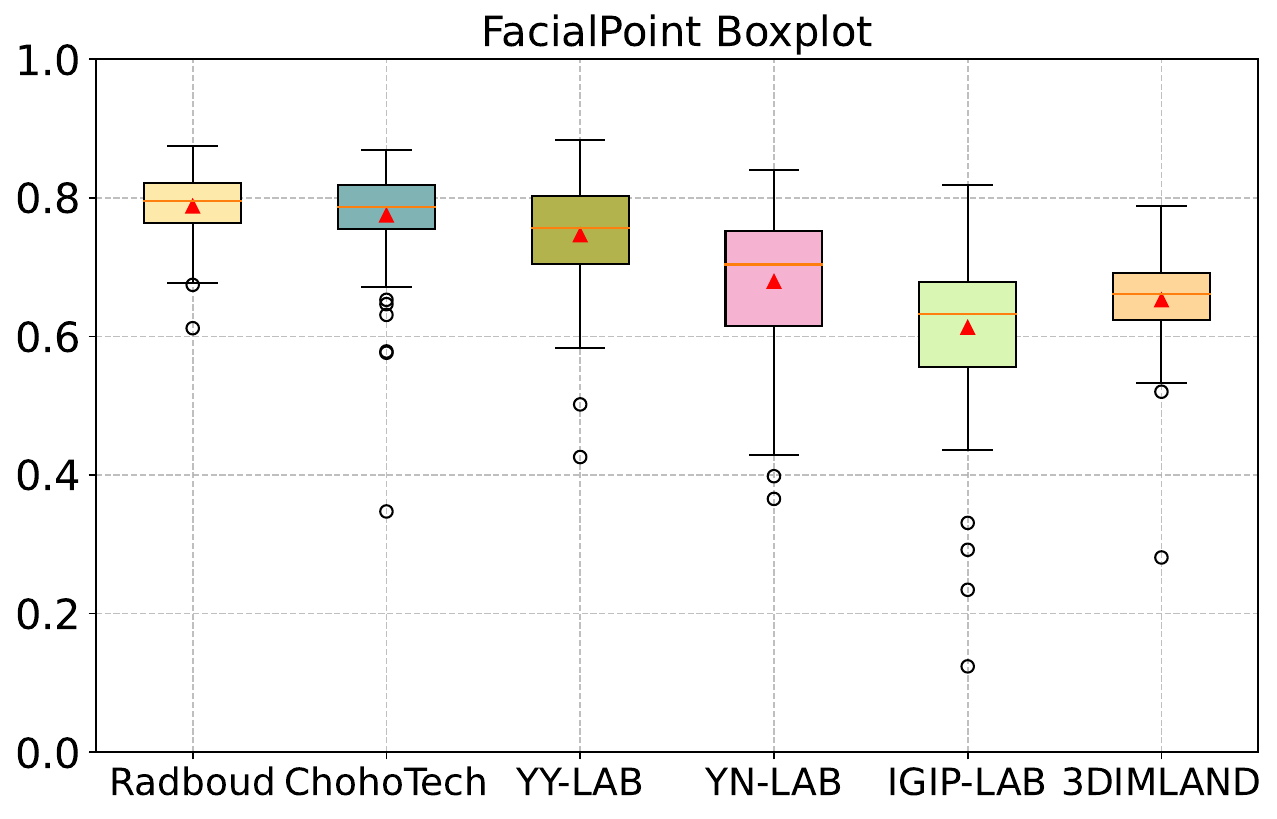}
                \caption{mAP for facial category}
                \label{fig:facialpoint_boxplot_mAP}
            \end{subfigure}

            \vspace{0.4cm} 

            \begin{subfigure}[t]{0.49\textwidth}
                \centering
                \includegraphics[width=\textwidth]{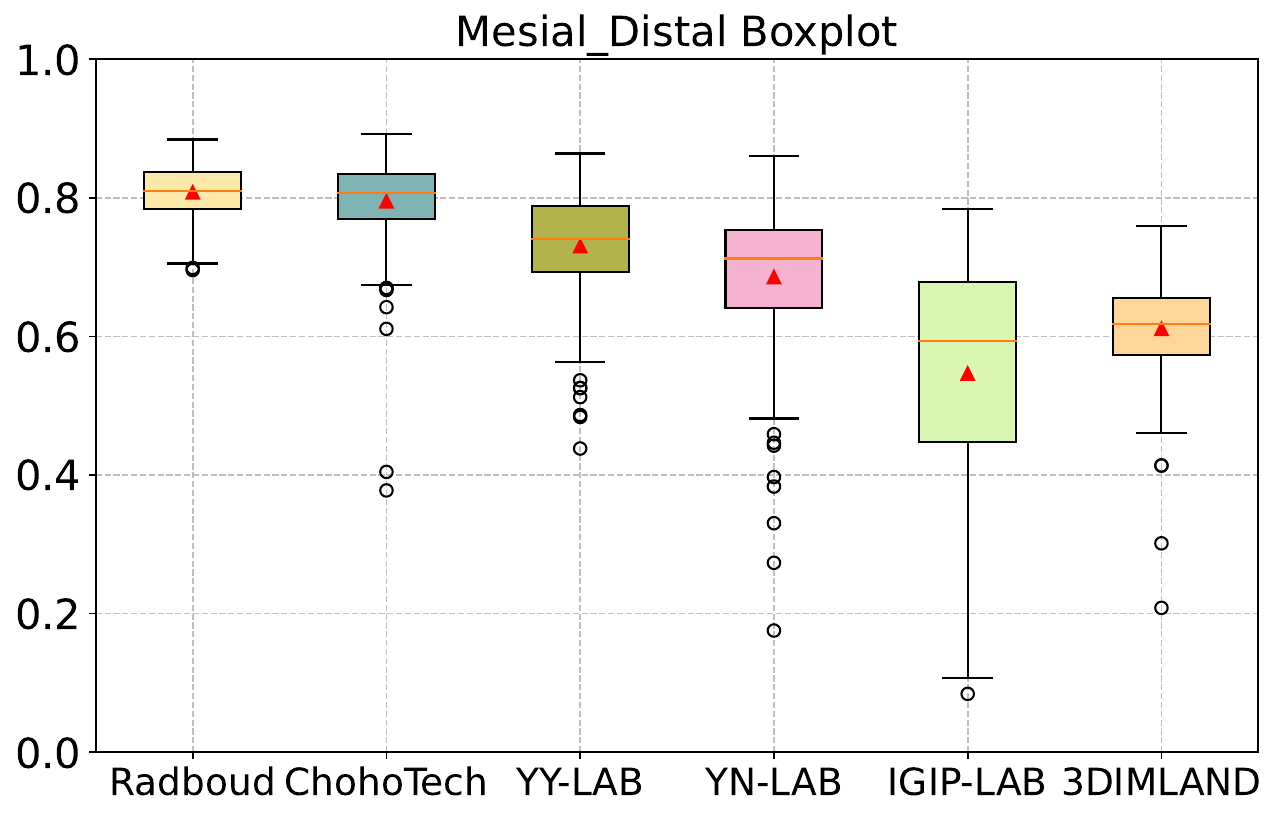}
                \caption{mAP for mesial\_distal category}
                \label{fig:mesial_boxplot_mAP}
            \end{subfigure}\hfill
            \begin{subfigure}[t]{0.49\textwidth}
                \centering
                \includegraphics[width=\textwidth]{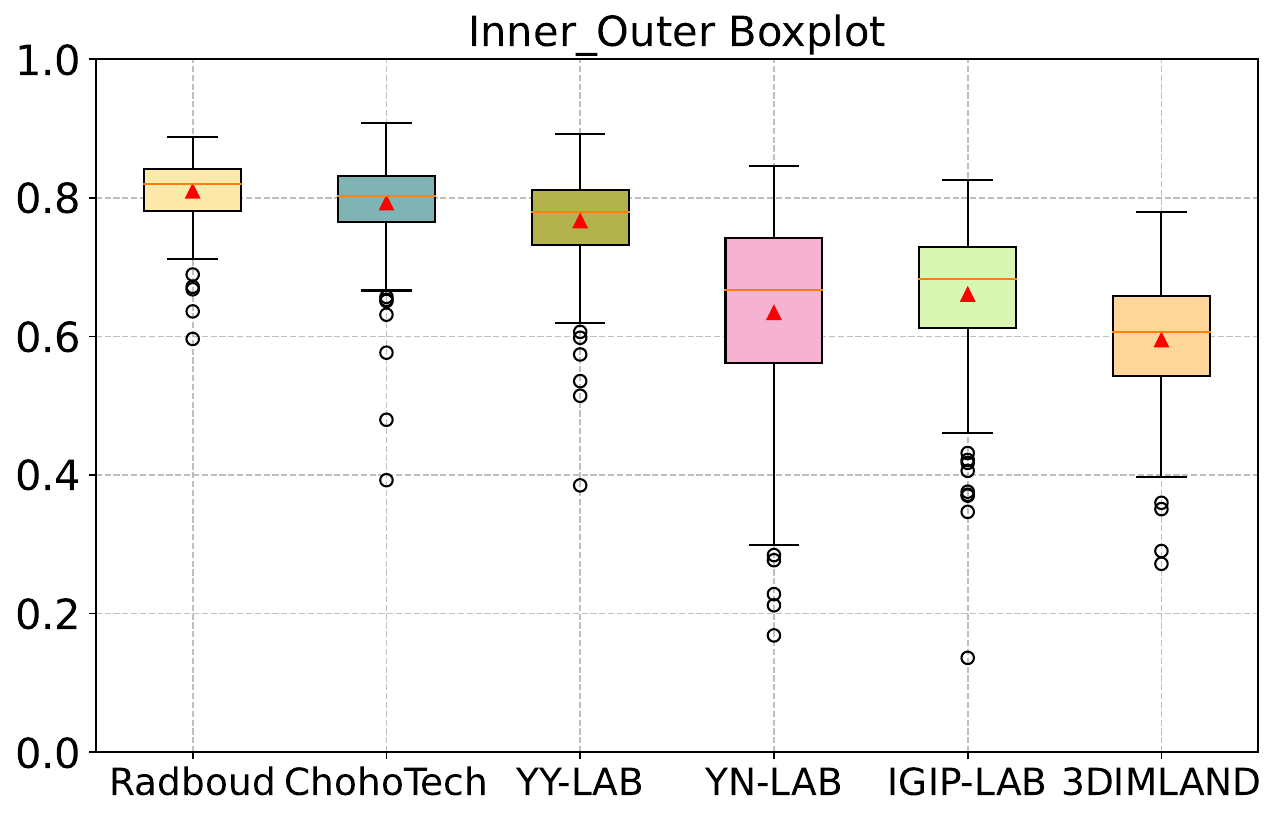}
                \caption{mAP for inner\_outer category}
                \label{fig:distal_boxplot_mAP}
            \end{subfigure}
        \end{minipage}
    \end{adjustbox}
    \caption{mAP score for each landmark category per scan.}
    \label{fig:mAP_boxplot}
\end{figure*}

\begin{figure*}[h]
    \centering
    \resizebox{0.8\textwidth}{!}{%
        \begin{minipage}{\textwidth}
            \begin{subfigure}[t]{0.49\textwidth}
                \centering
                \includegraphics[width=\textwidth]{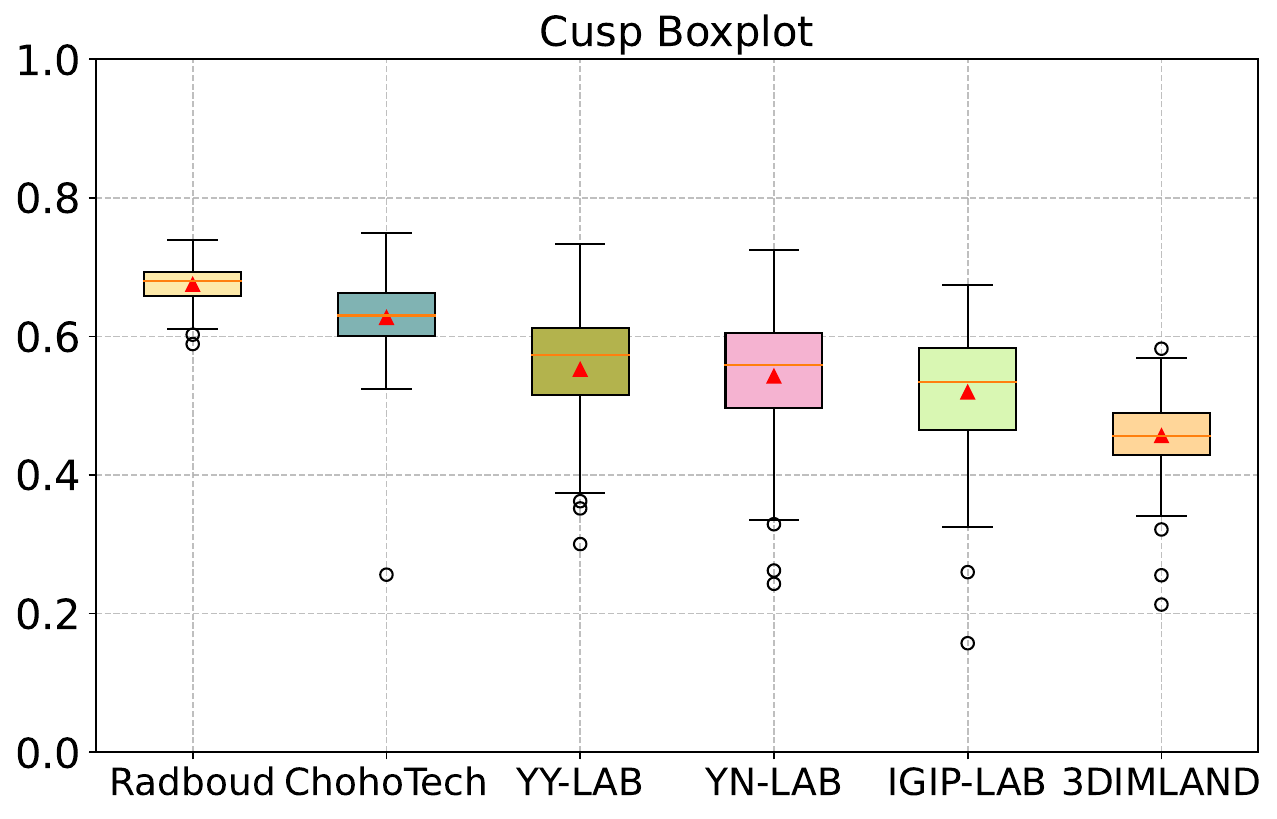}
                \caption{mAR for cusp category}
                \label{fig:cusp_boxplot_mAR}
            \end{subfigure}\hfill
            \begin{subfigure}[t]{0.49\textwidth}
                \centering
                \includegraphics[width=\textwidth]{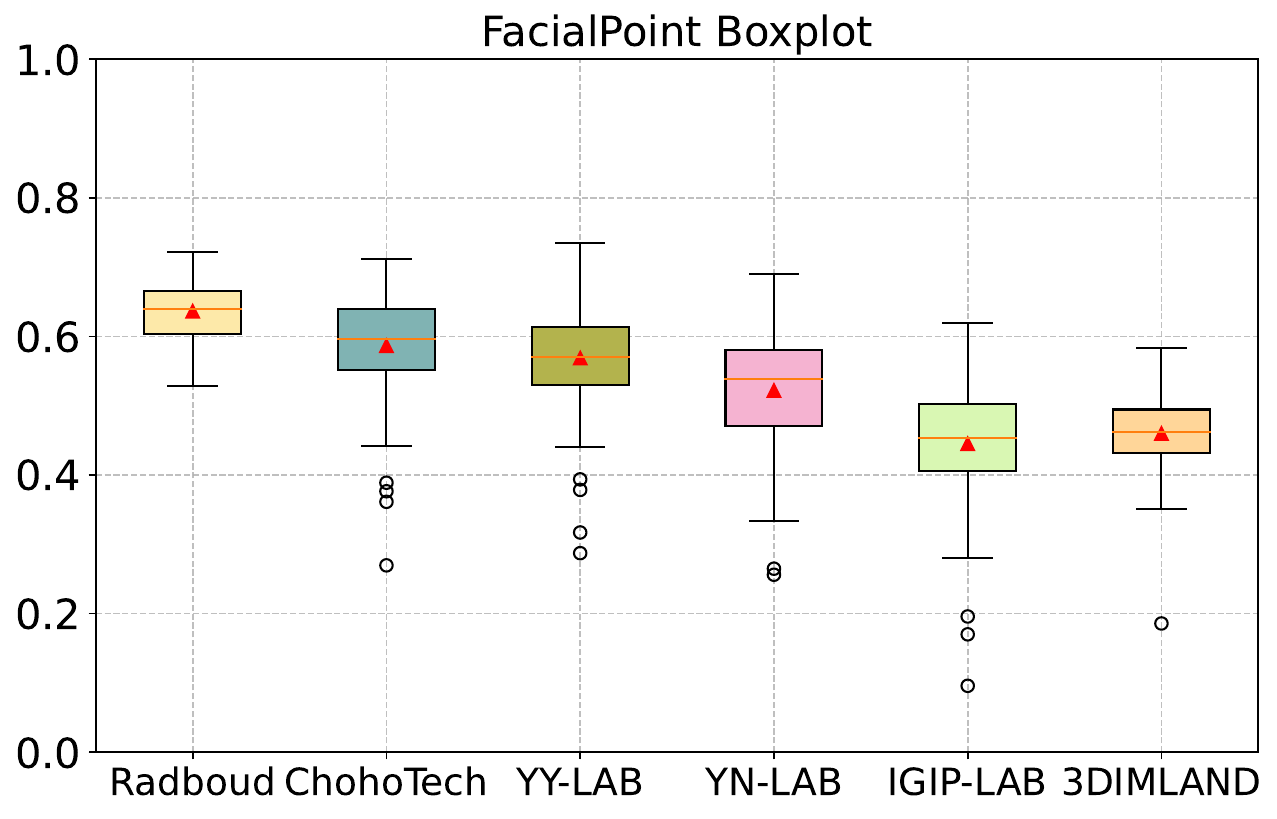}
                \caption{mAR for facial category}
                \label{fig:facialpoint_boxplot_mAR}
            \end{subfigure}

            \vspace{0.4cm} 

            \begin{subfigure}[t]{0.49\textwidth}
                \centering
                \includegraphics[width=\textwidth]{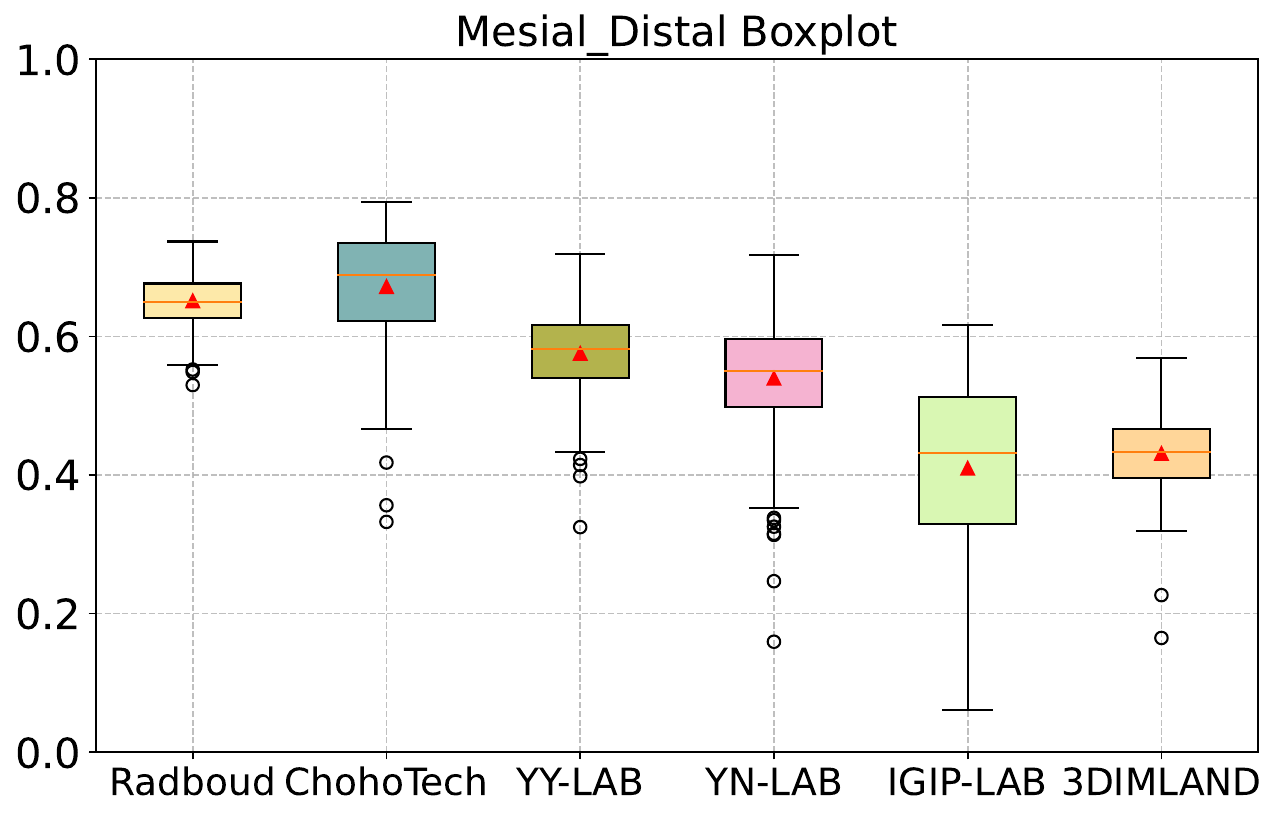}
                \caption{mAR for mesial\_distal category}
                \label{fig:mesial_distal_boxplot_mAR}
            \end{subfigure}\hfill
            \begin{subfigure}[t]{0.49\textwidth}
                \centering
                \includegraphics[width=\textwidth]{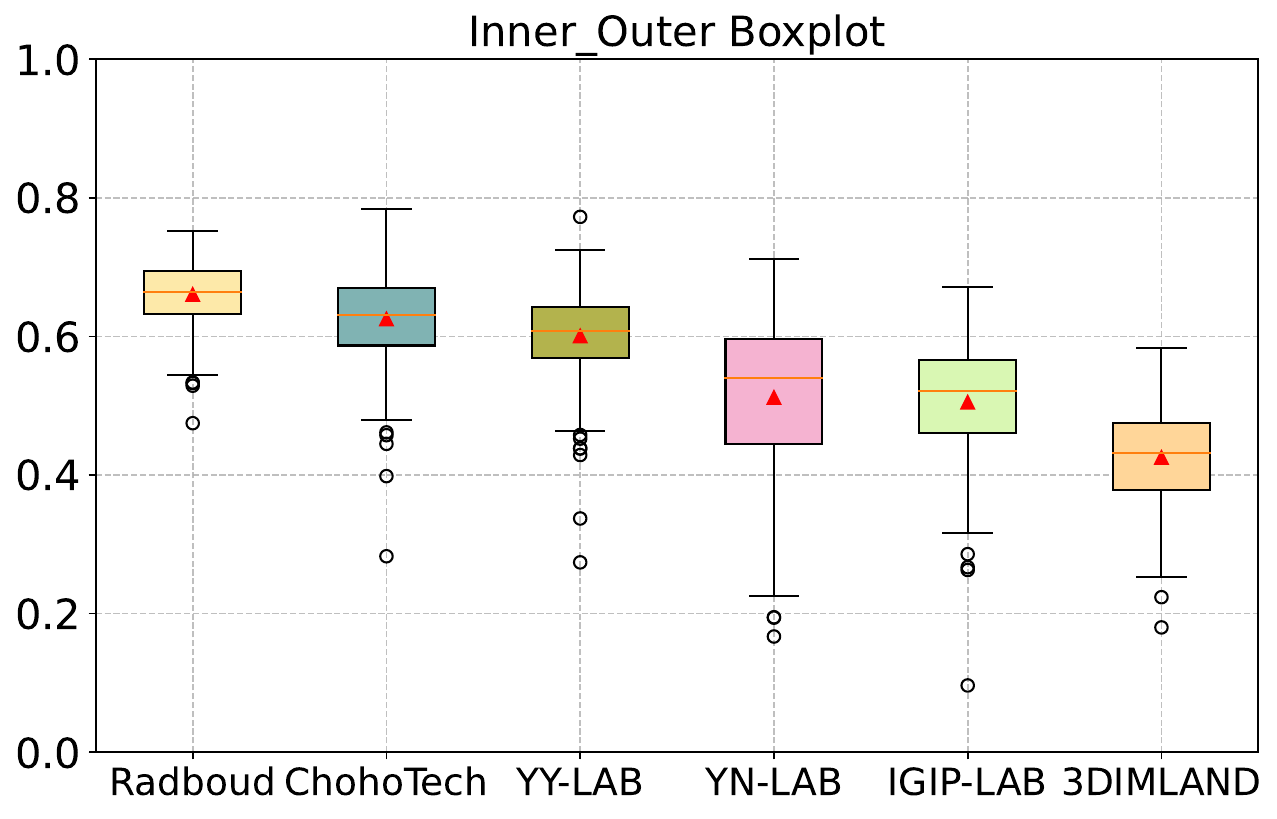}
                \caption{mAR for inner\_outer category}
                \label{fig:inner_outer_boxplot_mAR}
            \end{subfigure}
        \end{minipage}%
    }
    \caption{mAR score for each landmark category per scan.}
    \label{fig:mAR_boxplot}
\end{figure*}

\subsection{\rev{Overall evaluation and ranking results}}

The performance of each participating team was evaluated based on the proposed ranking protocol, as detailed in the previous section. \rev{Table \ref{tab:ar_ap_rank} reports the quantitative comparison of the participating teams in terms of AP and AR across the four anatomical categories (C, F, I/O, and M/D), together with the aggregated metrics (mAP, mAR) and the final ranking score (RS).} Radboud team (Niels van Nistelrooij \etal) achieved the highest Rank Score of 0.9172, with a corresponding mAP of 0.785 \rev{$\pm$ 0.04} and mAR of 0.656 \rev{$\pm$ 0.04}, indicating an outstanding balance between precision and recall. \rev{Importantly, the relatively small standard deviations highlight the stability of the method across scans.} On the other hand, the 3DIMLAND team (Tibor Kubík \etal) ranked lowest with a Rank Score of 0.0325, highlighting areas requiring significant improvement in landmark detection.

Radboud team (Niels van Nistelrooij \etal) and ChohoTech team (Huikai Wu) consistently achieved high mAP scores across all categories, with notable strengths in the Inner/Outer and Mesial/Distal categories, where their scores exceeded 0.78. \rev{However, the standard deviations reveal differences in robustness between the two top methods. While Radboud shows low variability (mAP 0.78 $\pm$ 0.04), ChohoTech exhibits larger dispersion (mAP 0.77 $\pm$ 0.07), indicating that its performance varies more across scans and may degrade on challenging cases.}

Teams like IGIP-LAB team (Weijie Liu \etal) and 3DIMLAND team (Tibor Kubík \etal) demonstrated lower mAP scores, particularly in the Mesial/Distal category, suggesting challenges in detecting contact points between teeth.

The highest mAR values were also observed in teams like Radboud team (Niels van Nistelrooij \etal) and ChohoTech team (Huikai Wu), with Radboud team excelling in the Inner/Outer category, achieving a score of 0.661 \rev{$\pm$ 0.05}, and performing equally well in the Mesial/Distal category with a score of 0.651 \rev{$\pm$ 0.04}. The ChohoTech team demonstrated strong recall performance in the Mesial/Distal category, achieving a score of 0.672 \rev{$\pm$ 0.08}.  Lower-ranked teams like the IGIP-LAB team (Weijie Liu \etal) and 3DIMLAND team (Tibor Kubík \etal) struggled once again to achieve recall rates comparable to those of higher-performing teams across all landmark categories. \rev{This is compounded by their relatively large standard deviations, which reveal high variability across scans.}

\figuresabvr \ref{fig:mAP_boxplot} and \ref{fig:mAR_boxplot} provide an analysis of the distribution of the values for mAP and mAR metrics across scans and categories, respectively. From the boxplots of mAP and mAR across different categories, it is evident that the Radboud team (Niels van Nistelrooij \etal) demonstrates the most consistent performance, with narrow boxes and high median values for both metrics. This consistency shows that their model reliably performs well across various scans, exhibiting stable results. In addition, the ChohoTech team (Huikai Wu) shows a performance very similar to Radboud, with only a high median value in the Mesial/Distal category for the mAR metric.

In contrast, teams such as 3DIMLAND and IGIP-LAB exhibit greater variability in their performance, as indicated by wider boxes. This shows that their models' performance is inconsistent across scans, with some scans yielding high performance and others much lower, showing potential instability or difficulty in handling certain data types or conditions.

\subsection{\rev{Pairewise comparison results}}

\begin{figure}[t]
    \centering
    \revbox{
    \includegraphics[width=0.85\linewidth]{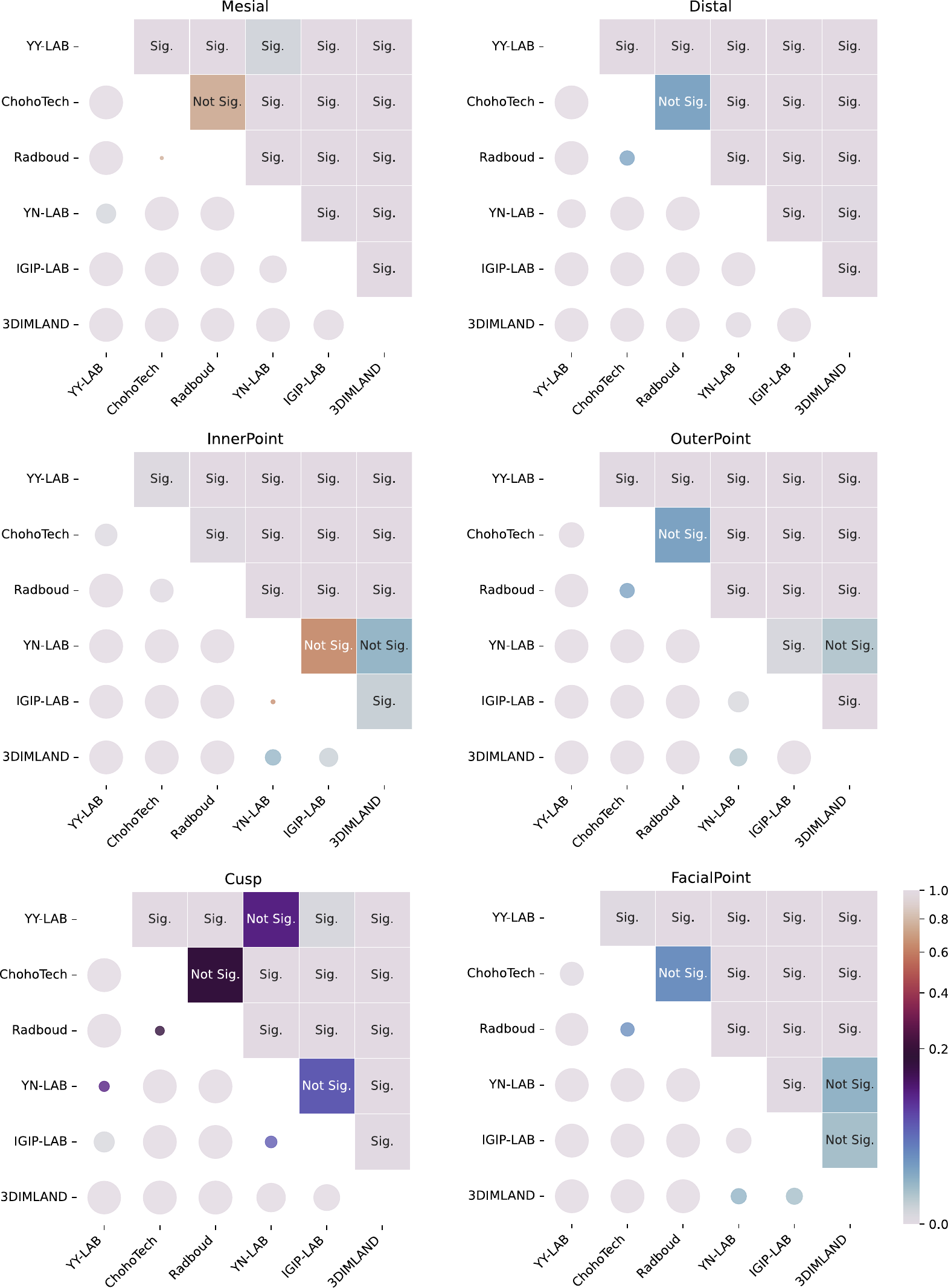}}
    \caption{\rev{Pairwise Wilcoxon signed-rank test \textit{p}-value heatmap for the mAP metric. Binary significance labels are shown in the upper triangle (``Sig" for $p < 0.001$; ``Not Sig" otherwise). The lower triangle represents \textit{p}-values in $[0,1]$ using discs. Larger discs indicate stronger statistical evidence of performance differences between teams (smaller $p$-values, closer to $0$), while smaller discs correspond to weaker or no evidence of differences (larger $p$-values).}}
    \label{fig:mAP_pairewise}
\end{figure}

\begin{figure}[t]
    \centering
    \revbox{
    \includegraphics[width=0.85\linewidth]{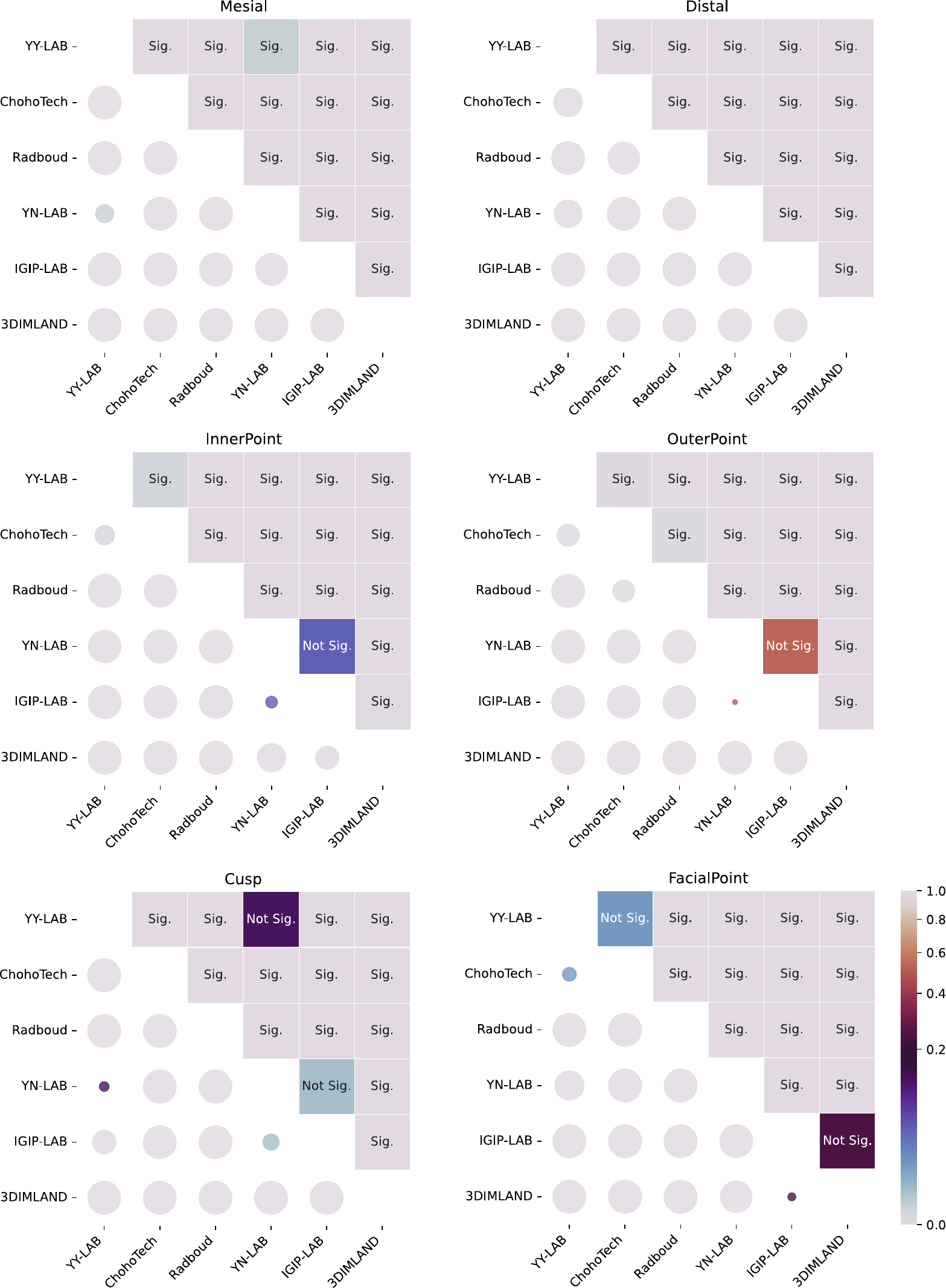}}
    \caption{\rev{Pairwise Wilcoxon signed-rank test \textit{p}-value heatmap for the mAR metric. Binary significance labels are shown in the upper triangle (``Sig." for $p < 0.001$; ``Not Sig." otherwise). The lower triangle represents \textit{p}-values in $[0,1]$ using discs. Larger discs indicate stronger statistical evidence of performance differences between teams (smaller $p$-values, closer to $0$), while smaller discs correspond to weaker or no evidence of differences (larger $p$-values).}}
    \label{fig:mAR_pairewise}
\end{figure}

\rev{\figuresabvr \ref{fig:mAP_pairewise}  and \ref{fig:mAR_pairewise} present the pairwise Wilcoxon signed-rank test results for the mAP and mAR metrics across all teams. Consistent with earlier findings, a clear separation is observed between the two top-performing methods (Radboud and ChohoTech) and the remaining teams. Their two approaches demonstrate statistically significant differences against most competitors, as reflected by multiple large discs in the lower triangle and consistent “Sig” labels in the upper triangle. In contrast, the comparison between the two best-performing models shows predominantly non-significant differences for both metrics. For example, for the mAP metric on the cusp landmark, the disk is small, which is consistent with the close numerical values reported in \tableabvr \ref{tab:ar_ap_rank} (0.77 $\pm$ 0.06 vs. 0.76 $\pm$ 0.07). These findings motivate a closer analysis of the design choices underlying the two top-ranked methods. }

\begin{figure*}[h]
    \centering
    \revbox{
    \resizebox{0.9\textwidth}{!}{%
        \begin{minipage}{\textwidth}

        \begin{subfigure}[t]{0.24\textwidth}
            \centering
            \includegraphics[width=\linewidth]{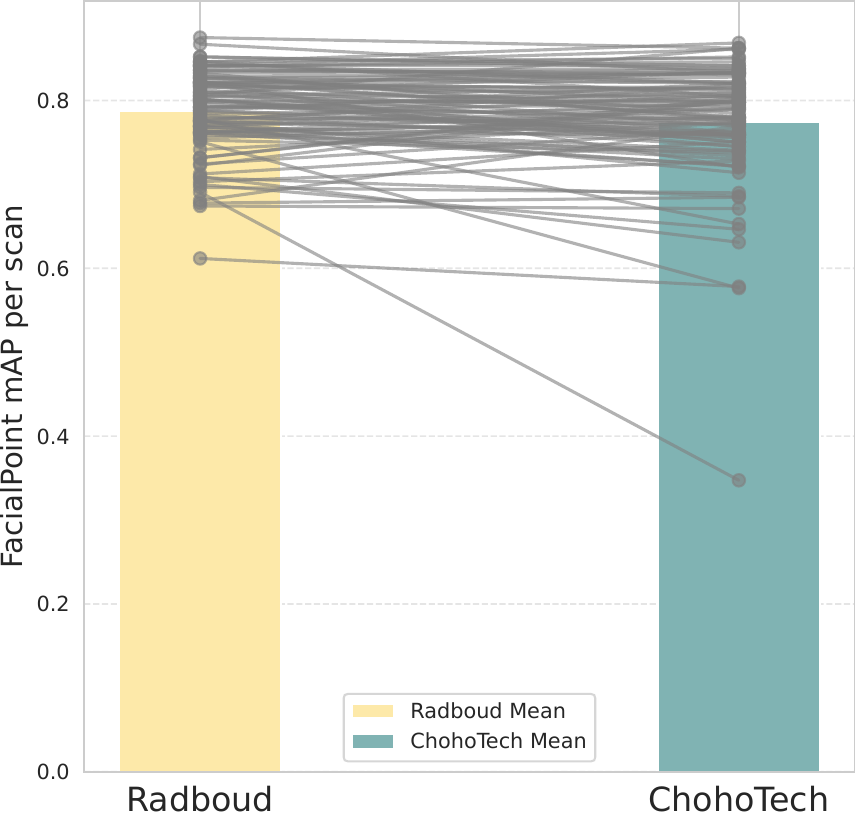}
            \caption{Facial}
        \end{subfigure}
        \hfill
        \begin{subfigure}[t]{0.24\textwidth}
            \centering
            \includegraphics[width=\linewidth]{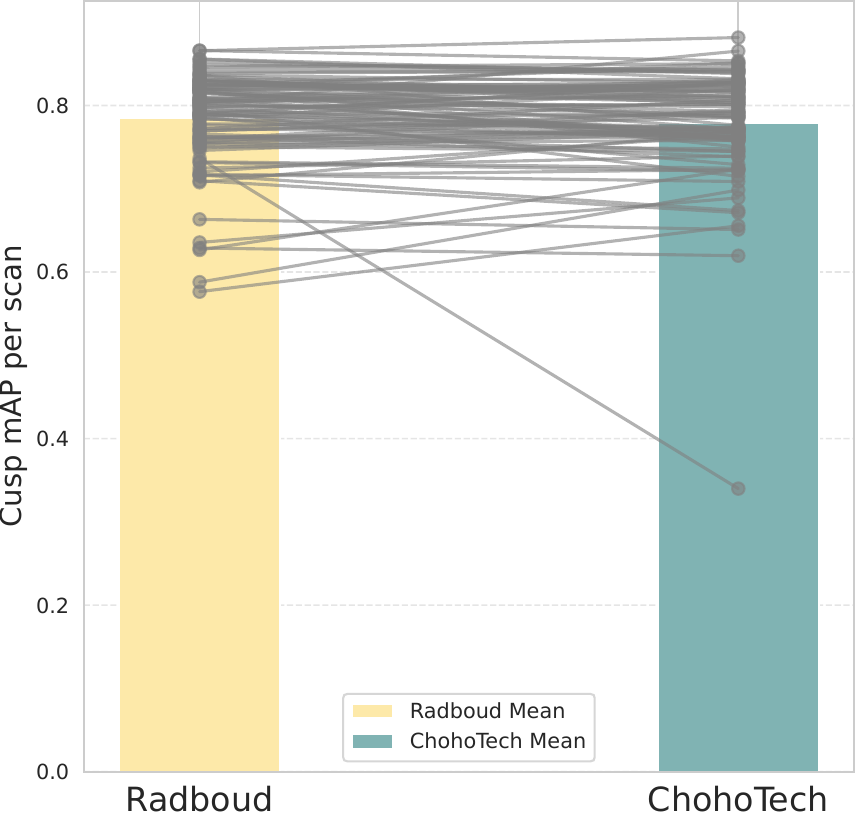}
            \caption{Cusp}
        \end{subfigure}
        \hfill
        \begin{subfigure}[t]{0.24\textwidth}
            \centering
            \includegraphics[width=\linewidth]{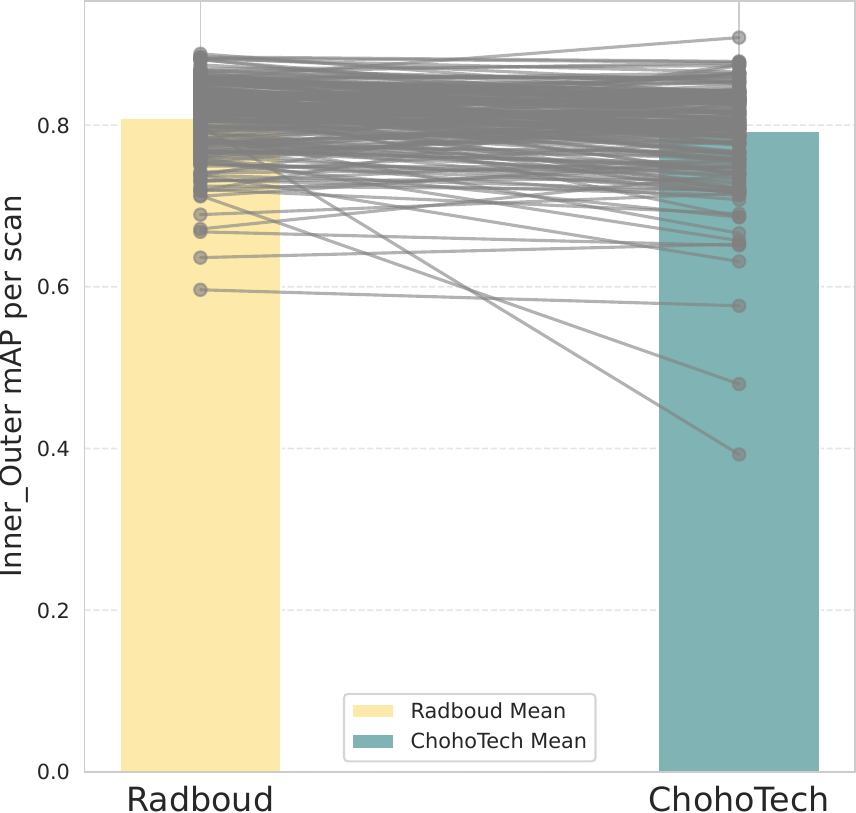}
            \caption{Inner/Outer}
        \end{subfigure}
        \hfill
        \begin{subfigure}[t]{0.24\textwidth}
            \centering
            \includegraphics[width=\linewidth]{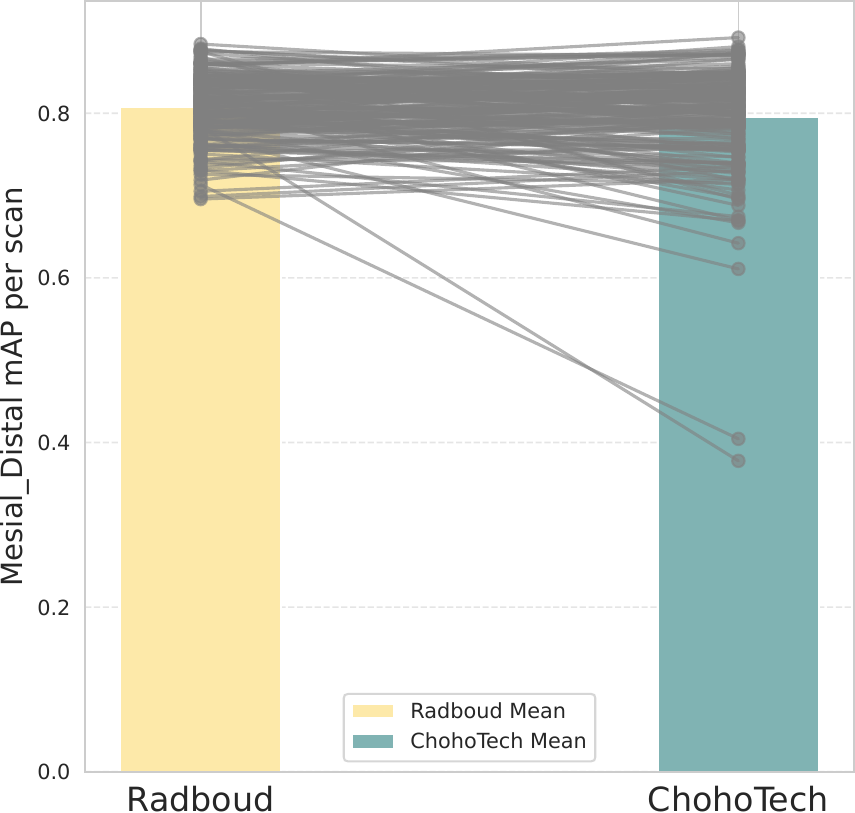}
            \caption{Mesial/Distal}
        \end{subfigure}

        \vspace{0.4cm}

        \begin{subfigure}[t]{0.24\textwidth}
            \centering
            \includegraphics[width=\linewidth]{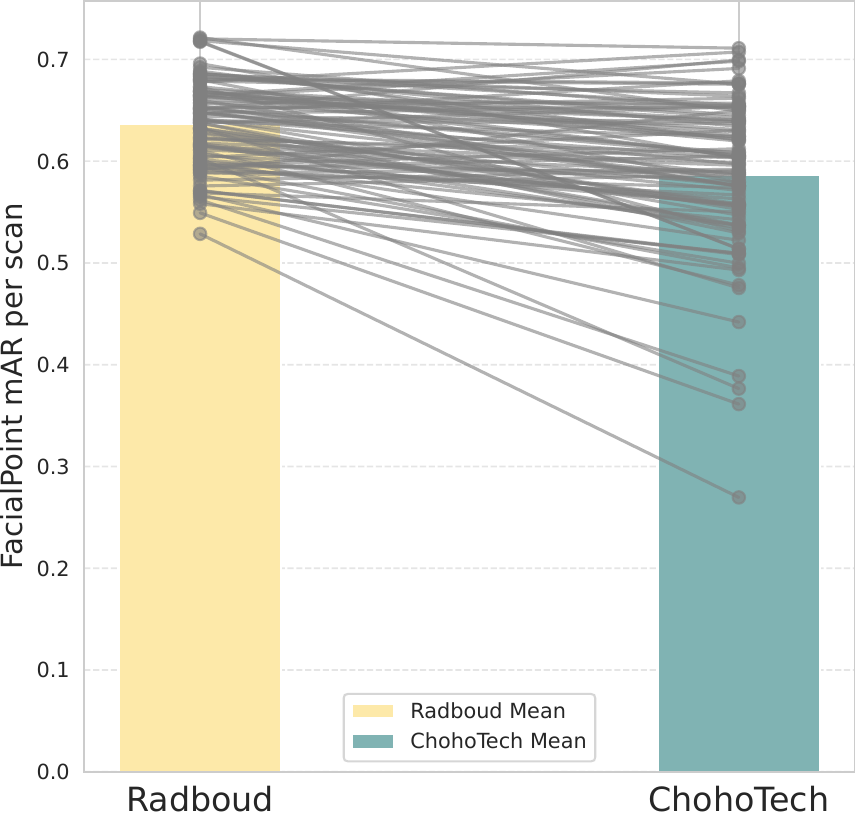}
            \caption{Facial}
        \end{subfigure}
        \hfill
        \begin{subfigure}[t]{0.24\textwidth}
            \centering
            \includegraphics[width=\linewidth]{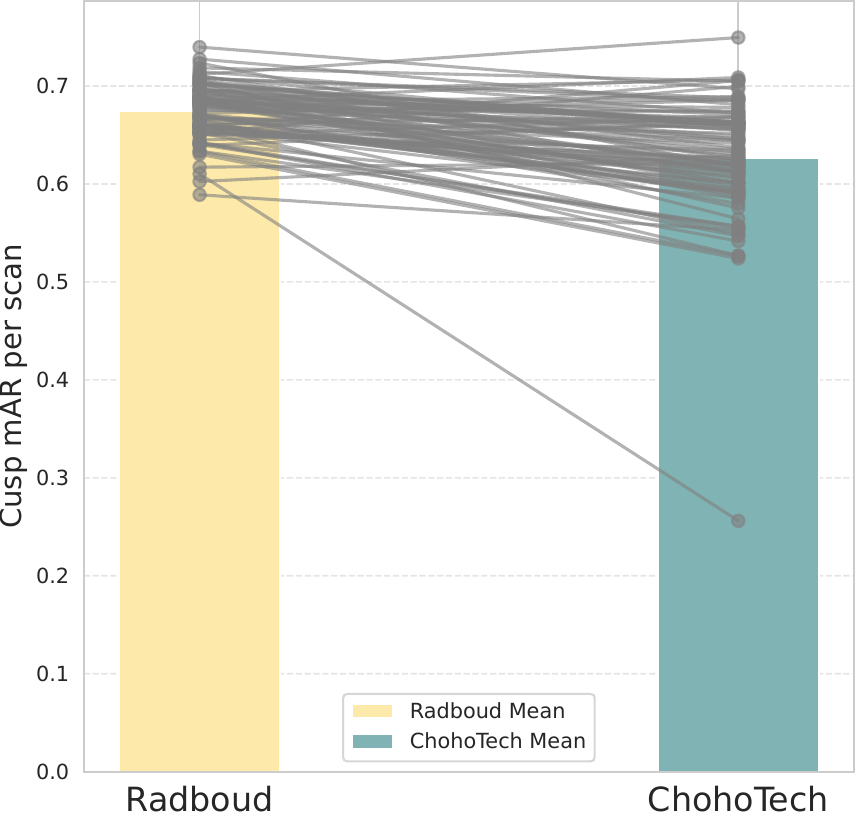}
            \caption{Cusp}
        \end{subfigure}
        \hfill
        \begin{subfigure}[t]{0.24\textwidth}
            \centering
            \includegraphics[width=\linewidth]{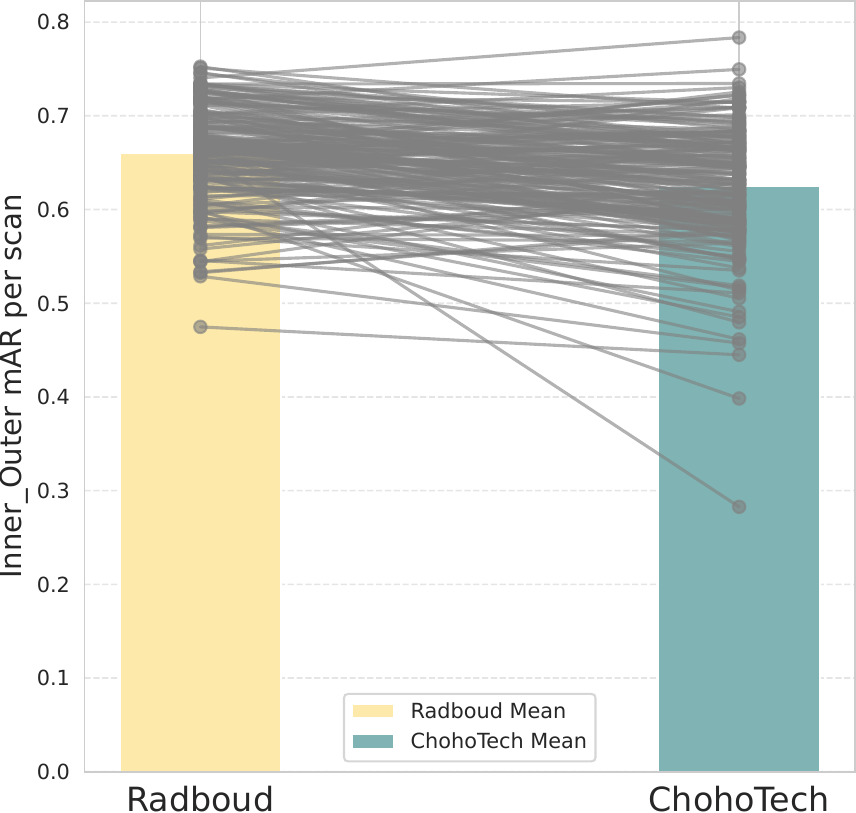}
            \caption{Inner/Outer}
        \end{subfigure}
        \hfill
        \begin{subfigure}[t]{0.24\textwidth}
            \centering
            \includegraphics[width=\linewidth]{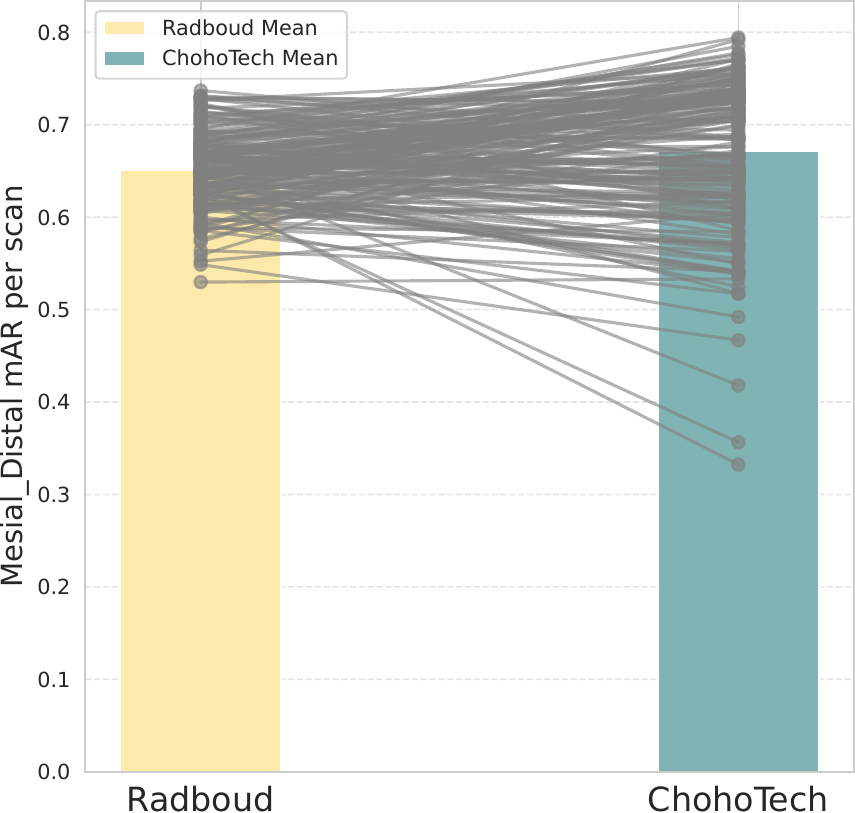}
            \caption{Mesial/Distal}
        \end{subfigure}

        \end{minipage}
    }}
    \caption{\rev{Paired comparison of mAP (top row) and mAR (bottom row) between the top two ranked teams (Radboud and ChohoTech) across different landmark categories. Each gray line corresponds to a single scan and connects the performance of the two methods, highlighting per-scan differences. Colored bars represent the mean performance across all scans.}}
    \label{fig:paired_comparison}
\end{figure*}

\rev{The Radboud and ChohoTech teams achieved top results using fundamentally different strategies. Radboud's two-stage, segmentation-based ToothInstanceNet first captures global context from a low-resolution full-arch scan for accurate tooth labeling, then focuses on high-resolution individual tooth crops ($\approx$10,000 points per tooth) to capture fine geometric detail. Separate decoders for segmentation and each landmark class enable anatomical specialization, while weighted DBSCAN refines landmark coordinates, contributing to high overall mAP across stricter thresholds. In contrast, ChohoTech's single-stage, segmentation-free ORNet relies on an offset regression branch to refine each point toward its nearest landmark, achieving sub-millimeter precision. By avoiding segmentation, it reduces complexity and error propagation, and high point density (up to 20,000 points per jaw) compensates for the lack of segmentation, producing an efficient and precise landmark localization pipeline. \figureabvr \ref{fig:paired_comparison} presents a paired per-scan comparison of mAP and mAR between the Radboud and ChohoTech teams. Most paired lines show a consistent downward trend from Radboud to ChohoTech across all landmark categories, indicating that the observed performance gap arises at the individual-scan level rather than being driven by a few outliers. Both methods tend to struggle on the same scans, indicating that cases difficult for Radboud are generally difficult for ChohoTech as well. However, ChohoTech exhibits a more pronounced performance drop on these challenging scans. This effect is especially pronounced for mAR, indicating that ChohoTech's performance gap arises mainly from missed detections rather than localization errors, particularly on complex scans. The superior robustness of the Radboud method can be attributed in part to its use of specialized decoders that independently propose landmarks for each anatomical class. This class-specific specialization allows the network to learn distinct features for different types of landmarks, improving detection rates on irregular morphologies and contributing to its higher recall. In addition, the use of high-resolution tooth crops allows the model to capture fine-grained local geometry, while the post-processing stage further refines the predicted landmark coordinates and improves spatial consistency. Together, these components lead to higher recall and more stable performance across scans. Overall, these results highlight the superior robustness of the Radboud method, which maintains more consistent detection performance across the dataset and is less sensitive to scan-level difficulties.}

\clearpage
\subsection{\rev{Qualitative results}}
\begin{figure}
    \centering
    \includegraphics[width=0.87\linewidth]{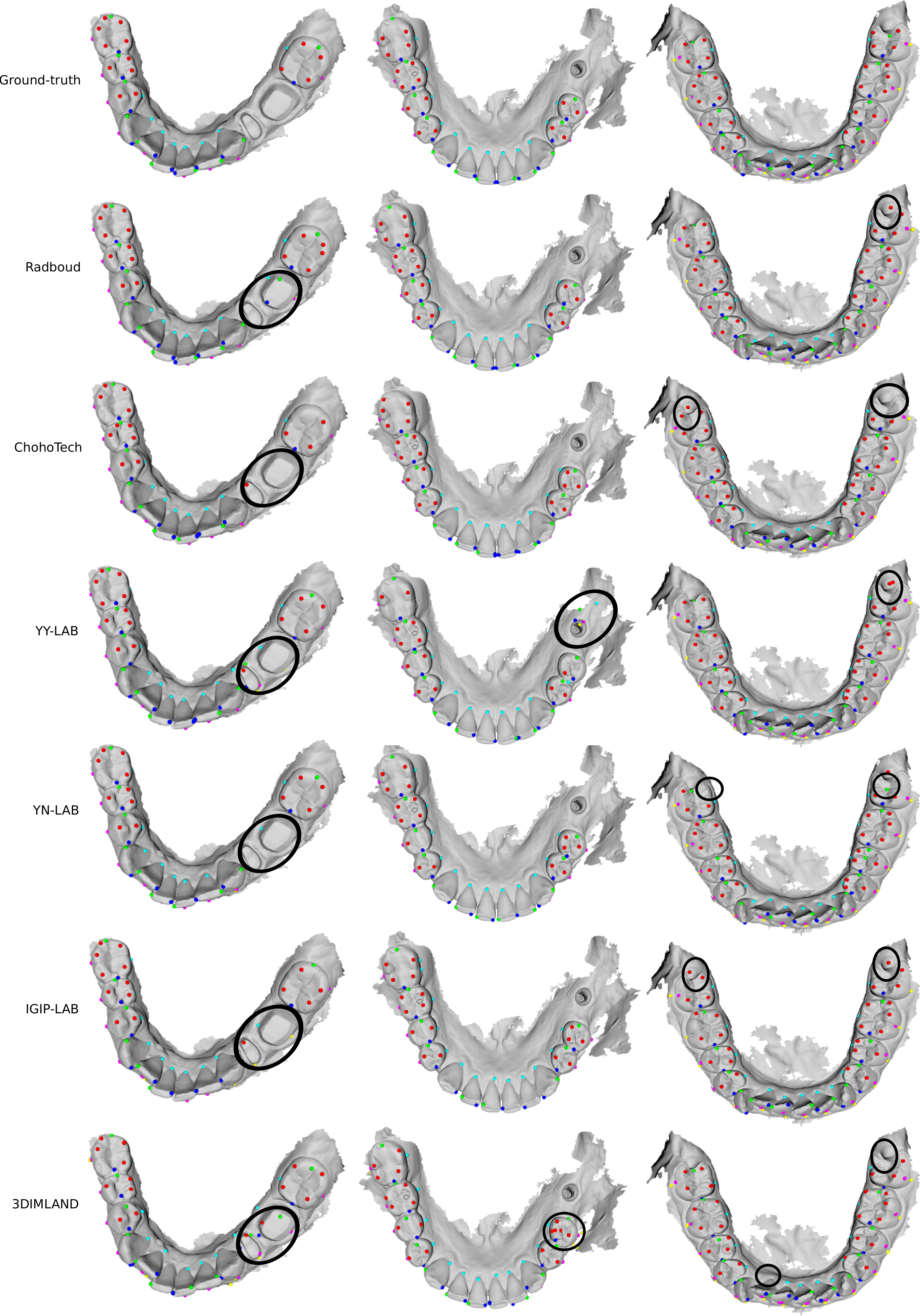}
    \caption{Overall visual comparison of landmark detection obtained by the six methods. Each column corresponds to an example of a lower jaw. Black ellipses denote examples of undetected landmarks or localization errors.}
    \label{fig:visual_comparison}
\end{figure}

\figureabvr \ref{fig:visual_comparison} provides an overall visual comparison of landmark detection obtained by the six methods. Each column represents a lower jaw example, allowing for a direct comparison of how well each team's method performs. Notably, the Radboud team's approach shows the best alignment between the detected landmarks and the ground truth, indicating a higher level of accuracy and consistency in landmark localization. On the other hand, the 3DIMLAND team's approach struggles with missing or mislocated landmarks.

\begin{figure}
    \centering
    \includegraphics[width=\linewidth]{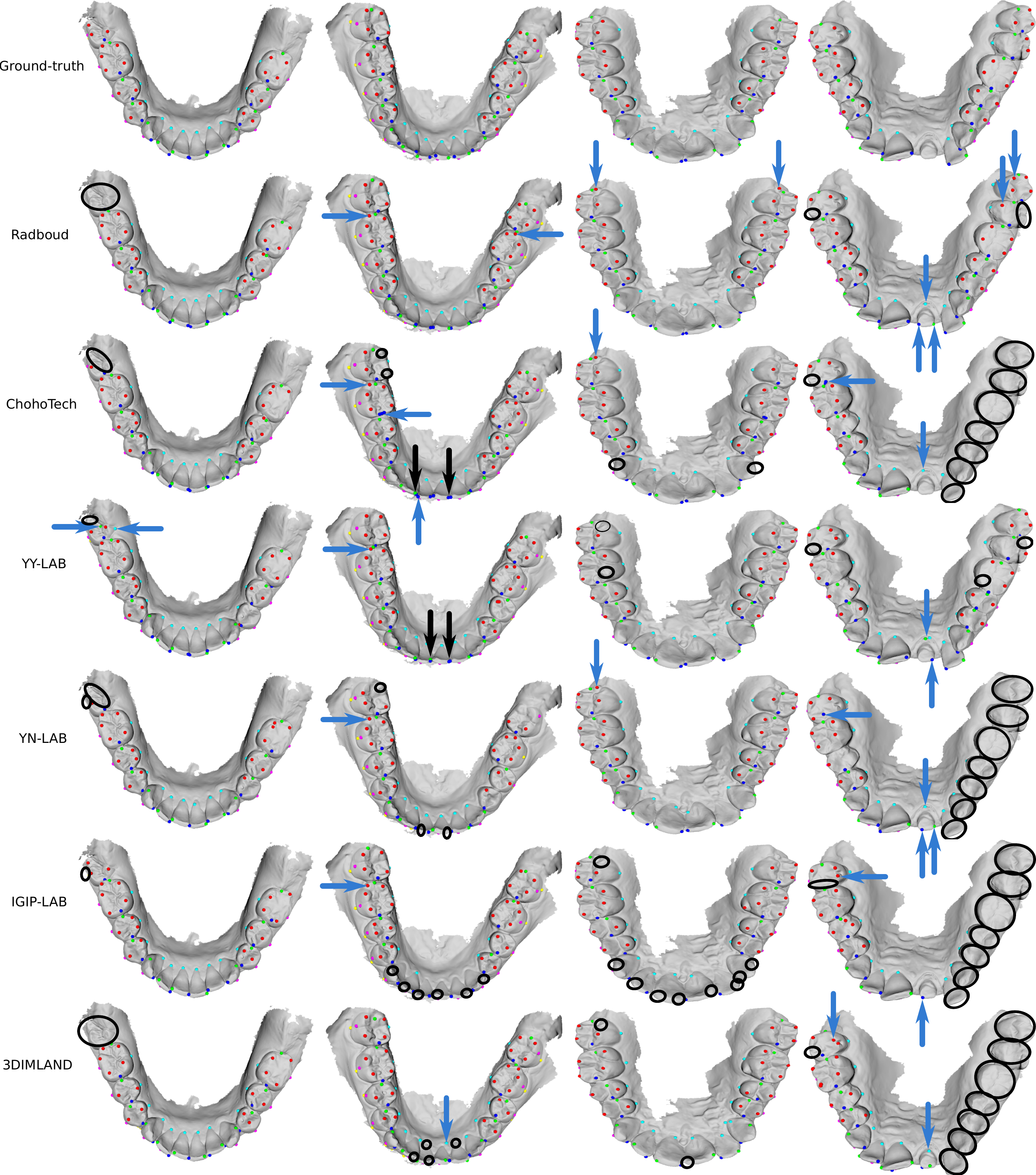}
    \caption{Visual comparison of hard cases for landmark detection obtained by the six methods. The first and second columns correspond to examples of a lower jaw. The third and fourth columns depict examples of an upper jaw. Black ellipses denote examples of undetected landmarks. Examples of errors in landmarks are marked with blue arrows, while category errors are marked with black arrows.}
    \label{fig:visual_comparison_hard_cases}
\end{figure}

\rev{\figureabvr \ref{fig:visual_comparison_hard_cases} presents difficult cases, in which all methods perform different mistakes. These cases exhibit missing teeth, severe arch asymmetry, pronounced tooth rotations, and incomplete surface regions, in which the proposed methods failed to detect all landmarks accurately.} In the final case (fourth case), none of the teams successfully identified all the ground-truth landmarks. The Radboud and YY-LAB teams missed some cusp points, while the ChohoTech, YN-LAB, IGIP-LAB, and 3DIMLAND teams failed to detect landmarks on the right half of the jaw. In the second case, the ChohoTech and YY-LAB teams misclassified some landmarks, with some distal points labeled as mesial and some mesial points labeled as distal.

\rev{In summary, the top-performing methods demonstrate robust performance, and in successful cases the achieved accuracy lies within human annotation variability and clinically acceptable limits, supporting their practical clinical applicability.}

\clearpage

\section{Discussion}
\rev{The challenge results highlight a clear performance gap between the top-performing teams and the lower-ranked ones. Radboud and ChohoTech consistently outperformed others in both precision and recall, demonstrating the effectiveness of their landmark detection strategies, while teams such as IGIP-LAB and 3DIMLAND struggled to achieve comparable results, revealing areas needing significant improvement. These differences cannot be explained solely by the choice of backbone, segmentation strategy, or data augmentation, but also by how each method handles local geometric detail and refinement. In the following, we discuss the observed trends and high-level features that appear most relevant to performance.}

\subsection{Runtime} The evaluation of these approaches can be expanded by considering factors such as model complexity, including computational resources, which affect real-world applicability. Additionally, time consumption is crucial for assessing the feasibility of these methods, particularly for large-scale or real-time tasks. 
\tableabvr \ref{tab:time_consuming} presents a comparative analysis of the total time taken and the average processing time per scan for different teams. Each team's approach varies significantly in terms of computational complexity. Although the Radboud team stands out as one of the most effective,  their approach has a relatively high total processing time (2125.3 seconds). Conversely, the 3DIMLAND and ChohoTech teams achieved the lowest processing times, with averages of 10.67 and 10.89 seconds per scan, respectively. Notably, the ChohoTech team achieved results comparable to the Radboud team's best model in terms of mAP and mAR, demonstrating that their approach balances efficiency and accuracy effectively. In contrast, while the 3DIMLAND team achieved the fastest processing time, their performance in mAP and mAR is lower.

\begin{table}[b]
\renewcommand{\arraystretch}{1.2}
\centering
\begin{tabular}{@{}lccc@{}}\toprule
\multicolumn{1}{c}{\textbf{Team}} &  \textbf{Total Time (sec)} & \textbf{Avg/scan (sec)} \\
\hline
\fcolorbox{white}{lightpink}{\rule{0pt}{5pt}\rule{5pt}{0pt}} Radboud team  (Niels van Nistelrooij \etal) &  2125.3 & 21.25 \\
\fcolorbox{white}{lightblue}{\rule{0pt}{5pt}\rule{5pt}{0pt}} ChohoTech team (Huikai Wu) & 1089.9 & 10.89 \\
\fcolorbox{white}{olive}{\rule{0pt}{5pt}\rule{5pt}{0pt}}  YY-LAB team  (Kaibo Shi \etal) & 2289.7 & 22.89 \\
\fcolorbox{white}{pink}{\rule{0pt}{5pt}\rule{5pt}{0pt}}  YN-LAB team (Zhu Xiaoying \etal) & 2962.3 & 29.62 \\
\fcolorbox{white}{lightgreen}{\rule{0pt}{5pt}\rule{5pt}{0pt}}  IGIP-LAB team (Weijie Liu \etal)  & 1313.5 & 13.13 \\
\fcolorbox{white}{rose}{\rule{0pt}{5pt}\rule{5pt}{0pt}} 3DIMLAND team (Tibor Kubík \etal)  & 1067.5 & 10.67 \\
\bottomrule
\end{tabular}
\caption{Time-consuming per team.}
\label{tab:time_consuming}
\end{table}

\subsection{\rev{Impact of tooth Segmentation}} The six approaches are divided into two categories: those that perform segmentation before landmark detection (\ie Radboud, YY-LAB, and YN-LAB teams) and those that do not involve tooth segmentation (\ie ChohoTech, IGIP-LAB, and 3DIMLAND teams). From \tableabvr \ref{tab:time_consuming}, we observe that not involving segmentation leads to a significant reduction in processing time. For instance, the ChohoTech team, which does not use tooth segmentation, achieved the second-best performance while also having one of the lowest processing times. This demonstrates that segmentation may not always be necessary for high accuracy, especially if the model is well-optimized. On the other side, the IGIP-LAB and 3DIMLAND teams also skipped the segmentation step, resulting in faster processing times but at the cost of reduced performance. This indicates that while removing segmentation improves computational efficiency, its impact on landmark detection accuracy largely depends on the overall effectiveness of the approach.

\subsection{\rev{Impact of global and local point sampling}}
\rev{A key factor could be the effective management of resolution and local point density, which directly impacts the precision of landmark detection. The first-ranked team, Radboud, leveraged segmentation to isolate individual teeth and allocate a very high resolution (around 10,000 points per tooth), thus capturing fine geometric structures essential for accurate landmark localization. In contrast, ChohoTech avoided segmentation but compensated by processing the full jaw with 20,000 points. An additional experiment made by the ChohoTech team demonstrated that increasing the point count from 5k to 20k nearly doubled detection precision, highlighting the importance of density even without explicit tooth separation. IGIP-LAB, despite using a relatively high number of points (16,384) across the entire jaw, still achieved a lower effective local resolution than both Radboud's per-tooth strategy and ChohoTech's denser sampling, which likely limited its ability to capture subtle geometric cues. Similarly, 3DIMLAND processed the full jaw at an even higher total resolution (64,000 points). However, this global allocation diluted the local density per tooth due to the inclusion of the gingiva and surrounding structures. As acknowledged by the 3DIMLAND team, this insufficient local detail resulted in a ``blurred" output from the distance decoder, hindering precise landmark extraction. Overall, high local point density on individual teeth is necessary but not sufficient for accurate landmark detection, as performance also depends on modeling choices and post-processing strategies.}

\subsection{\rev{Role of the post-processing}}
\rev{Post-processing strategies also played a decisive role, particularly in handling challenging cases such as anatomical ``crowding". IGIP-LAB explicitly reported limitations of their no-segmentation approach in the anterior region, where landmarks such as mesial points are located in very close proximity. In these cases, their density-based clustering strategy struggled to properly separate neighboring landmarks, resulting in ambiguous or incorrect detections. Similar challenges were observed with the 3DIMLAND team, where crowded anatomical regions led to multiple modes of error, including imprecise localization of inner points near irregular tooth structures and missed detections of closely spaced landmarks. In contrast, higher-ranked teams introduced more robust refinement mechanisms tailored to such scenarios. Radboud addressed this issue using a Weighted DBSCAN clustering approach, where points were weighted inversely to their predicted distances.  This weighting strategy helped stabilize cluster centers and improve separation in dense regions. Similarly, ChohoTech applied class-specific Non-Maximum Suppression (NMS) after predicting offsets, effectively eliminating redundant detections and preserving only the most confident landmark candidates. These targeted post-processing techniques significantly improved robustness in crowded anatomical areas, contributing to better overall performance.}

\subsection{\rev{Limitations and future directions}}
\rev{The challenge provides a valuable benchmark while also revealing insights on the limitations and opportunities for future research.}
\rev{\paragraph{Dataset diversity} The proposed dataset provides a valuable resource for dental landmark detection and represents an important step toward advancing research in this area. While it already captures a meaningful range of clinical scenarios, future extensions could further enrich the dataset by incorporating additional edge cases and a wider spectrum of dental treatments and subfields.}

\rev{\paragraph{Model architectures} The proposed methods rely on similar paradigms, including transformers and point-based networks, which are trendy architectures that perform very well. However, this architectural bias naturally encourages the exploration of alternative approaches, such as implicit or hybrid representations, which could be better suited for certain specific cases.}

\rev{\paragraph{Future directions} Enriching the proposed dataset is an essential step for advancing future research. To further enhance its representativeness of diverse clinical settings, future versions could include a broader range of variations, encompassing a wider demographic spectrum and various dental conditions. Additionally, incorporating more challenging real-world conditions, such as varying levels of noise and incomplete scans, would better prepare algorithms for the complexities encountered in clinical environments. This would contribute to improving the reliability and generalization of landmark detection models. Self-supervised learning could be explored to leverage large amounts of unlabeled intraoral scans, reducing reliance on manual annotations and offering strong potential for learning more robust and generalizable representations.}

\section{Conclusion}
\label{sec:conclusion}
The 3DTeethLand challenge, organized in collaboration with the International Conference on Medical Image Computing and Computer-Assisted Intervention (MICCAI) in 2024, focused on the development of algorithms for detecting teeth landmarks from intraoral 3D scans. This challenge marked the introduction of the first publicly available dataset for 3D teeth landmark detection, providing a crucial resource to foster community engagement in this vital area with considerable clinical implications.

Out of a total of 49 teams that officially participated in the 3DTeethLand challenge, six landmark detection algorithms were evaluated. This paper offers a comprehensive overview of the challenge setup, summarizes the algorithms submitted by the participating teams, and provides a detailed comparison of their performance.

During the final testing phase of the 3DTeethLand challenge, the algorithms were thoroughly evaluated on their ability to accurately detect teeth landmarks across a variety of intraoral 3D scans. The performance of each algorithm was assessed using key metrics, including the mAP and the mAR, which provided insights into both the accuracy and consistency of landmark detection. The top-performing teams consistently achieved high mAP and mAR values, indicating robust precision and recall across different scans and categories. Notably, the Radboud and ChohoTech teams exhibited superior performance. In contrast, other algorithms, such as those from the IGIP-LAB and 3DIMLAND teams, ranked lowest, needing improvement for a more stable and reliable landmark detection across diverse intraoral scans.

\section*{Acknowledgements}
This work was primarily supported by the Digital Research Center of Sfax, Tunisia. 
\section*{Appendix}
This section presents additional technical details related to the challenge that are not included in the main text but are essential for a complete and transparent description.

\begin{figure}
    \centering
    \revbox{
    \includegraphics[width=0.95\linewidth]{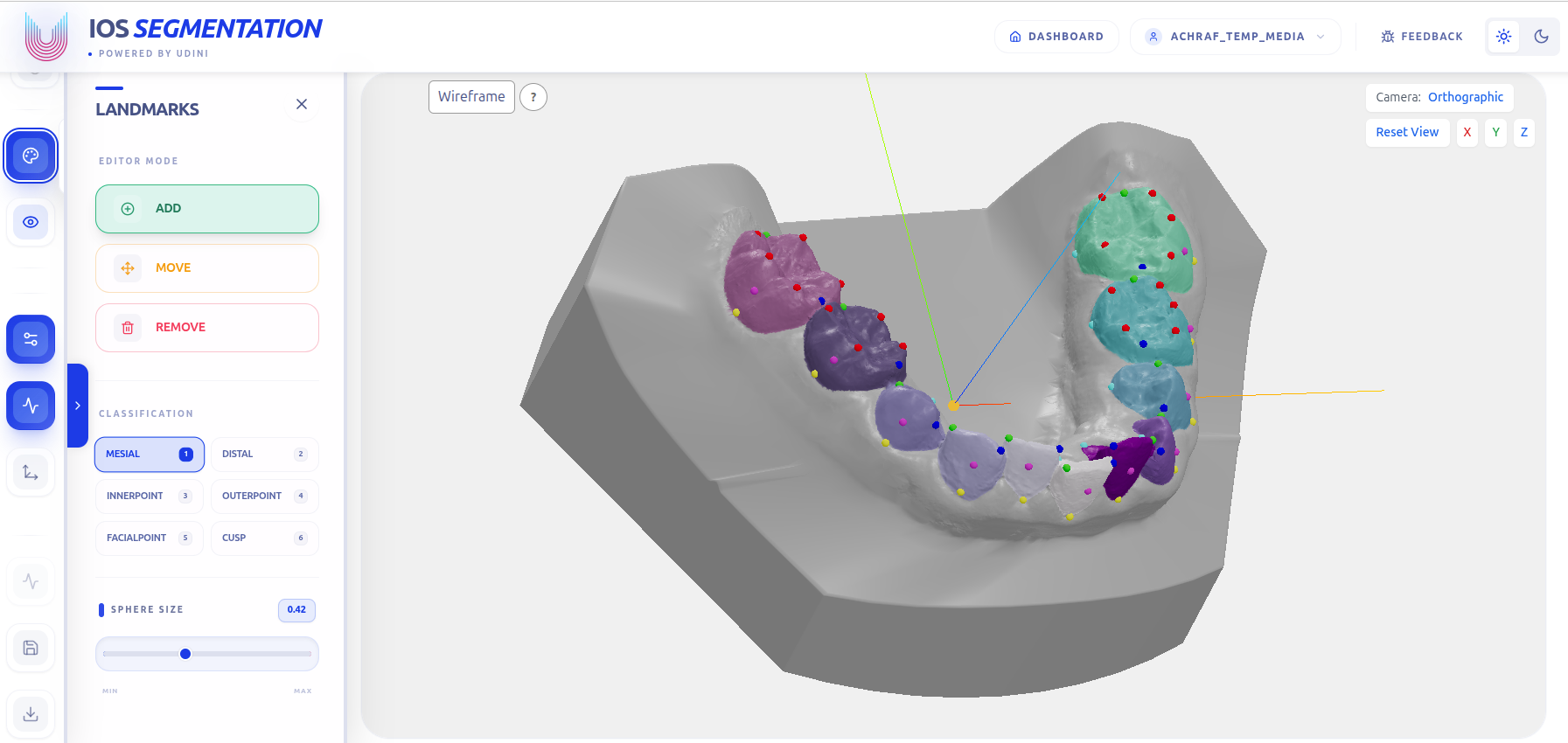}
    }
    \caption{\rev{Main user interface of the annotation tool (version 0.4.7) illustrating the dental landmark annotation functionalities, including landmark addition, adjustment (movement), and removal.}}
    \label{fig:main_interface_annotation_tool}
\end{figure}

\noindent\rev{\textbf{Annotation tool:} \figureabvr \ref{fig:main_interface_annotation_tool} presents the main user interface of the annotation tool, illustrating the dental landmark annotation functionalities, including landmark addition, adjustment (movement), and removal.}

\begin{figure}[h]
\centering
\revbox{
\includegraphics[width=0.7\textwidth]{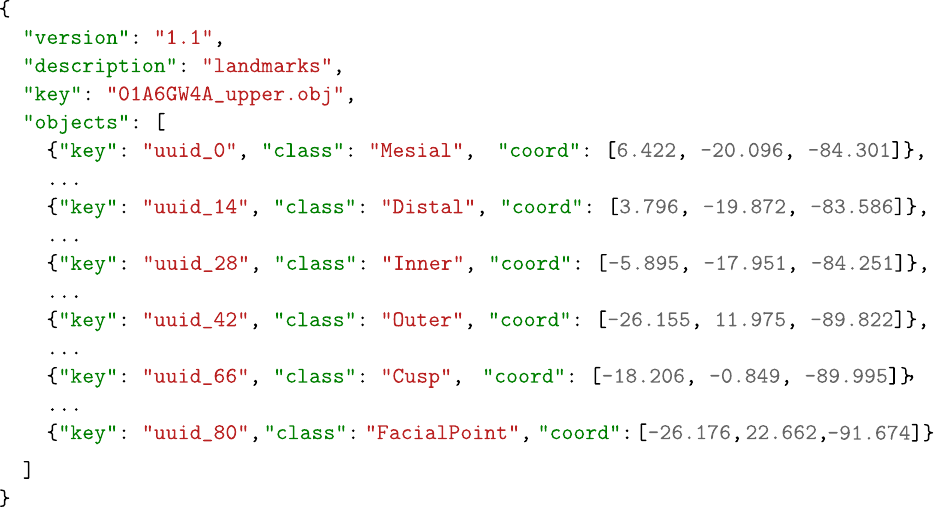}
}
\caption{\rev{Example of the file structure of landmarks annotation.}}
\label{fig:record}
\end{figure}

\noindent\rev{\textbf{Data structure}: \figureabvr \ref{fig:record} presents an example of landmarks file format.}

\noindent\textbf{Timeline:} Registration opened on the 1st of May, 2024, allowing teams to enroll and prepare for participation. The first batch of training data was released on the 27th of May, 2024, followed by the second batch on the 15th of June, 2024, enabling participants to begin data exploration and model development. The public test set became available on the 30th of July, 2024, at which time the submission portal for public test evaluation also opened.
The Final Phase began on the 26th of September, 2024, with the opening of the evaluation pipeline for final submissions. Teams were required to upload their Docker containers by the submission deadline on the 2nd of October, 2024. Algorithm descriptions and related papers were shared on the 29th of September, 2024, as part of the challenge documentation process. The challenge concluded on the 6th of October, 2024, during the MICCAI 2024 event.

\noindent\textbf{Organization:}
3DTeethLand is the successor to the 3DTeethSeg challenge, held as a recurring annual event with a fixed submission deadline. All evaluated algorithms are required to operate in a fully automatic manner, without any user interaction.

\noindent\textbf{Publication:} Throughout the challenge, participants had access to the GitHub repository \footnote{\href{https://github.com/crns-smartvision/3DTeethLand/tree/master}{https://github.com/crns-smartvision/3DTeethLand/tree/master}}, where they could find the source code for the evaluation script as well as a Docker template containing a baseline algorithm to be replaced for participation. All teams were required to share their code with the organizers for evaluation, and this step was mandatory for the top-ranking methods.

Each member of the team is listed as an author for the corresponding submission. This paper provides an overview of the challenge results and key proposals, including all contributors as co-authors. Participants are allowed to publish their own results independently after an embargo time of 6 months after the MICCAI event.

\rev{\noindent\textbf{Code availability:} \tableabvr \ref{tab:teams_refs_sources} regroups the reproducible resources of the participating teams, with direct links to their code Git repositories.}

\begin{table}[htbp]
\centering
\renewcommand{\arraystretch}{1.6}
\caption{Reproducible resources of participating teams.}
\label{tab:teams_refs_sources}

\resizebox{\textwidth}{!}{%
\rev{
\begin{tabular}{ll}
\toprule
\multicolumn{1}{c}{Team/Ref} & \multicolumn{1}{c}{Code availability} \\
\midrule
\fcolorbox{white}{lightpink}{\rule{0pt}{5pt}\rule{5pt}{0pt}}\quad  Radboud (Niels van Nistelrooij \etal) & \href{https://github.com/nnistelrooij/3dteethland/tree/final_test_phase}{\textcolor{blue}{\nolinkurl{https://github.com/nnistelrooij/3dteethland/tree/final_test_phase}}} \\
\fcolorbox{white}{olive}{\rule{0pt}{5pt}\rule{5pt}{0pt}}\quad YY-LAB (Kaibo Shi \etal) & \href{https://github.com/bibi547/TL-DETR}{\textcolor{blue}{\nolinkurl{https://github.com/bibi547/TL-DETR}}} \\
\fcolorbox{white}{pink}{\rule{0pt}{5pt}\rule{5pt}{0pt}}\quad YN-LAB (Zhu Xiaoying \etal) & \href{https://gitlab.com/m26409021/ynlab}{\textcolor{blue}{\nolinkurl{https://gitlab.com/m26409021/ynlab}}} \\
\fcolorbox{white}{lightgreen}{\rule{0pt}{5pt}\rule{5pt}{0pt}}\quad   IGIP-LAB (Weijie Liu \etal) & \href{https://github.com/weijiezaibenpao/igip-sdu-code}{\textcolor{blue}{\nolinkurl{https://github.com/weijiezaibenpao/igip-sdu-code}}} \\
 \fcolorbox{white}{lightblue}{\rule{0pt}{5pt}\rule{5pt}{0pt}}\quad ChohoTech (Huikai Wu) & \href{https://github.com/Choho-Tech-Wu/3DTeethLand}{\textcolor{blue}{\nolinkurl{https://github.com/Choho-Tech-Wu/3DTeethLand}}}  \\
 \fcolorbox{white}{rose}{\rule{0pt}{5pt}\rule{5pt}{0pt}}\quad 3DIMLAND (Tibor Kubik \etal) & \href{https://github.com/tescangroup/PTv3-for-detecting-anatomical-landmarks-in-dentistry}{\textcolor{blue}{\nolinkurl{https://github.com/tescangroup/PTv3-for-detecting-anatomical-landmarks-in-dentistry}}} \\
\bottomrule
\end{tabular}
}
}
\end{table}


\noindent\textbf{Author contributions:}
\textit{Conceptualization}: Achraf Ben-Hamadou, Nour Neifar, Ahmed Rekik, Oussama Smaoui, Sergi Pujades.
\textit{Methodology}: Achraf Ben-Hamadou, Nour Neifar, Ahmed Rekik, Oussama Smaoui, Firas Bouzguenda, Sergi Pujades.
\textit{Evaluation Script}:  Oussama Smaoui.
\textit{Software challenge Submission}: Niels van Nistelrooij, Shankeeth Vinayahalingam, Kaibo Shi, Hairong Jin, Youyi Zheng, Tibor Kubík, Oldřich Kodym, Petr Šilling, Kateřina Trávníčková, Tomáš Mojžiš, Jan Matula, Xiaoying Zhu, Jeffry Hartanto, Kim-Ngan Nguyen, Tudor Dascalu, Huikai Wu, Weijie Liu, Shaojie Zhuang, Guangshun Wei, Yuanfeng Zhou.
\textit{Validation}: Achraf Ben-Hamadou, Nour Neifar, Ahmed Rekik, Oussama Smaoui.
\textit{Writing, Review and Editing}: Achraf Ben-Hamadou, Nour Neifar, Ahmed Rekik, Oussama Smaoui, Sergi Pujades.

\noindent\textbf{Conflicts of interest:} There are no conflicts of interest. Udini is the principal sponsor of the challenge by collecting and providing clinical data with the contribution of the Digital Research Center of Sfax. Only the organizers and members of their immediate team have access to test case labels.

 \bibliographystyle{elsarticle-num}
 \bibliography{main}

\end{document}